\documentclass[letterpaper]{article} 
\usepackage{aaai2027} 
\usepackage[hyphens]{url}  
\usepackage{graphicx} 
\usepackage{natbib}  
\usepackage{caption} 
\usepackage{algorithm}
\usepackage{algorithmic}
\usepackage{amsmath}
\usepackage{amssymb}
\usepackage{booktabs}
\usepackage{multirow}
\usepackage{graphicx}
\usepackage{xcolor}
\usepackage{colortbl}

\definecolor{lightgray}{RGB}{235,235,235}
\definecolor{lightblue}{RGB}{225,239,255}

\providecommand{\NA}{\textemdash}

\usepackage{newfloat}
\usepackage{listings}
\DeclareCaptionStyle{ruled}{labelfont=normalfont,labelsep=colon,strut=off} 
\floatstyle{ruled}
\newfloat{listing}{tb}{lst}{}
\floatname{listing}{Listing}

\usepackage{booktabs}

\title{A Model Merging Approach for Continual MLLM Unlearning}

\author{
Yuhang Wang\textsuperscript{1},
Linlin Zhang\textsuperscript{1},
Haoxuan Ji\textsuperscript{2},
Xianmin Ye\textsuperscript{1},
Zhenxing Niu\textsuperscript{1},
Haichang Gao\textsuperscript{1}
}

\affiliations{
\textsuperscript{1}Xidian University, Xi'an, China\\
\textsuperscript{2}Xi'an Jiaotong University, Xi'an, China\\
}

\begin{document}

\maketitle

\begin{abstract}
Multimodal large language model (MLLM) unlearning methods have been proposed to remove private, sensitive, or proprietary information from well-trained models. 
However, most existing MLLM unlearning methods are designed for one-shot requests and fail to adequately address continual scenarios, as repeatedly applying one-shot operations leads to cumulative utility degradation, unlearning rebound, and retention drift. 
We introduce \textbf{\emph{Merging for Continual Unlearning}} (\textbf{MCU}), an approach that \emph{dynamically} merges multiple one-shot unlearning adapters into a unified adapter upon receiving each new unlearning request.
Through a leave-one-out merging analysis, we reveal that these unlearning adapters exhibit strong cross-task dependencies. Such dependencies have two contrasting effects: they can facilitate cross-task unlearning transferability, but they can also introduce severe interference that degrades unlearning effectiveness and compromises retained knowledge.
To address this challenge, MCU projects the adapters into a shared representation space, preserves their dominant directions, suppresses over-concentrated coordinates, and reconfigures cross-task dependencies to mitigate interference while enhancing transferability.
Experiments on ICU-Bench and MLLMU-Bench demonstrate that MCU achieves superior unlearning effectiveness while preserving both retained knowledge and general multimodal utility.
\vspace{-0.8em}
\end{abstract}

\vspace{-0.5em}
\section{Introduction}
\begin{figure}[t]
    \centering
    \includegraphics[width=0.90\linewidth]{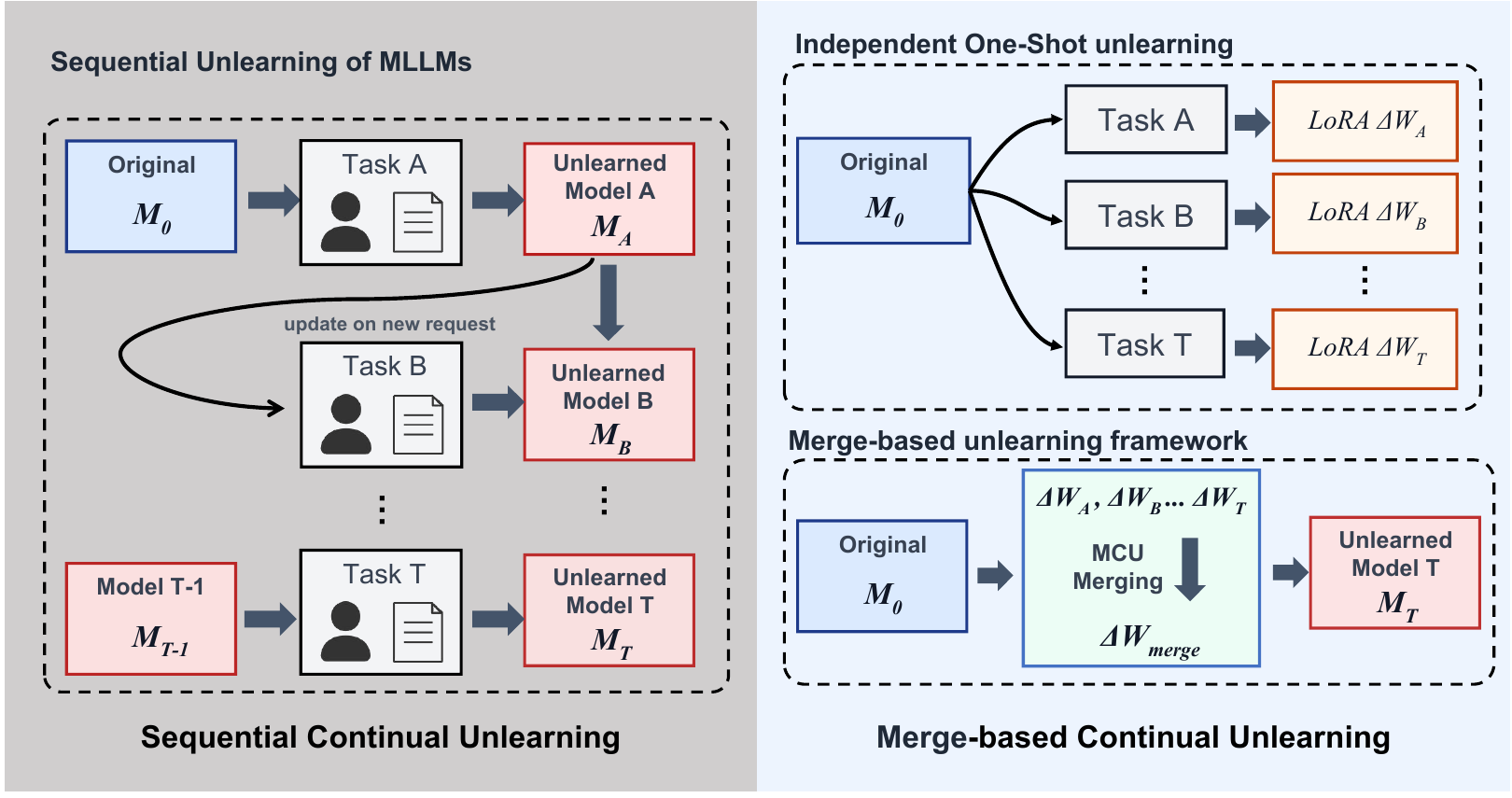}
    \vspace{-0.8em}
    \caption{Comparison between sequential continual unlearning and the proposed MCU. Existing methods repeatedly update the current model, coupling new requests to preceding model states and accumulating unlearning rebound, retention drift, and utility degradation.}
    \label{fig1}
    
    \vspace{-1.8em}
\end{figure}

Multimodal large language models (MLLMs) may encode private, proprietary, copyrighted, or outdated multimodal information from their training data~\cite{pi2024mllm,li2024digger,cohen2025performance}. 
Machine unlearning seeks to remove specified information while preserving unrelated capabilities~\cite{garg2020formalizing,gupta2021adaptive,sekhari2021remember}.
In practice, deletion requests often arrive sequentially, and existing continual unlearning methods typically process each incoming request by applying a new one-shot model or adapter update to the current model ~\cite{gao2025large,shi2025muse,kawakami2025pulse}.
As illustrated in Fig.~\ref{fig1}, this sequential paradigm couples each request to preceding model states, which can overwrite earlier unlearning effects and accumulate parameter drift, resulting in unlearning rebound, retention drift, and utility degradation ~\cite{li2026mlubench,wang2026icu}.
The key challenge is therefore to accommodate an expanding sequence of deletion requests while preserving both historical unlearning effects and general multimodal capabilities.

To address this challenge, we formulate continual multimodal unlearning as a model-merging problem. Specifically, we regard each task-specific unlearning adapter as a task vector and achieve continual unlearning by merging these task vectors into a unified adapter. Since all one-shot adapters are independently derived from the same base model, their updates can be represented and jointly processed in a shared parameter space without repeatedly modifying the current model. Through a leave-one-out merging study, we observe that the knowledge associated with a specific target remains partially forgotten even when its corresponding adapter is excluded from the merge, as shown in Fig.~\ref{fig3}. This indicates that adapters trained for other requests can also contribute to forgetting that target, revealing shared or overlapping unlearning directions across tasks. We refer to this phenomenon as \emph{cross-task unlearning transfer}.

Recent studies~\cite{lin2026cata} show that unlearning adapters exhibit structures distinct from ordinary capability adaptation, often concentrating their effects within a limited set of parameters or update directions.
To characterize the structures underlying cross-task unlearning transfer, we analyze the task singular directions of different adapters in a shared space, following structural analyses of model merging~\cite{gargiulo2025task}.
We find substantial dependencies among their dominant singular directions. These dependencies may be synergistic, reinforcing unlearning across requests, or interfering, weakening target unlearning or damaging retained knowledge.
Under direct merging, synergistic dependencies can transfer and
reinforce unlearning effects across requests, whereas interference can
cancel target updates or broaden suppression to retained knowledge.
Indiscriminately removing all cross-task dependencies is also
undesirable because it may eliminate useful synergy.
Continual unlearning through adapter merging therefore poses three
coupled objectives: preserving retained knowledge, exploiting
cross-task synergy, and reducing interference.


To address this challenge, we propose \textbf{\emph{Merging for Continual Unlearning}} (\textbf{MCU}), a model-merging framework that consolidates accumulated one-shot unlearning adapters into a unified model update. We treat each adapter-induced update as an unlearning task vector and jointly merge the vectors derived from the same base model. MCU first maps these task vectors into a shared space, where it preserves their dominant directions and controls over-concentrated coordinates. It then performs dependency-aware direction reconfiguration to suppress antagonistic cross-task interactions while retaining beneficial shared structures that support cross-task unlearning transfer. Finally, the reconfigured task vectors are merged and reconstructed as a unified adapter.
Extensive experiments demonstrate that MCU enables one-shot unlearning adapters to support long-horizon continual unlearning, reducing unlearning rebound and retention drift while preserving effective unlearning and retained knowledge.
Our contributions are summarized as follows:
\begin{itemize}
    \item We introduce a merging-based framework for continual unlearning that replaces repeated model modification with the dynamic merging of one-shot unlearning adapters, and uncover \emph{cross-task unlearning transfer} among these adapters.


    \item We develop MCU, which preserves dominant task structures, controls over-concentrated coordinates, and reconfigures cross-task dependencies in a shared space to suppress antagonistic interactions while preserving beneficial unlearning transfer.

    \item Extensive experiments on ICU-Bench and MLLMU-Bench demonstrate that MCU enables one-shot unlearning adapters to support long-horizon continual unlearning, achieving superior unlearning effectiveness while preserving retained knowledge.

\end{itemize}

\vspace{-1.2em}
\section{Related Work}

\paragraph{Machine Unlearning.}
Machine unlearning removes designated training influence without
full retraining.
Representative approximate approaches use gradient-based objectives,
distribution-preserving regularization, or preference optimization
~\cite{thudi2022unrolling,liu2022continual,maini2024tofu,
rafailov2023direct,zhang2024negative}.
Recent work extends unlearning to MLLMs by editing modality-specific
neurons, visual modules, or cross-modal pathways
~\cite{huo2025mmunlearner,liu2025modality,li2024single,
wang2026null,wang2025mllm}.
Static MLLM unlearning is evaluated by benchmarks such as MU-Bench,
PEBench, MLLMU-Bench, CLEAR, UMU-Bench, and ForgetMe
~\cite{cheng2024mu,xu2025pebench,liu2025protecting,
dontsov2025clear,wang2026umu,yu2025forgetme}, while MLUBench and
ICU-Bench study sequential deletion requests
~\cite{li2026mlubench,wang2026icu}.
Unlike methods that update the model separately or sequentially for
each request, MCU merges independently obtained unlearning adapters
and explicitly models their cross-request dependencies.

\vspace{-0.5em}
\paragraph{Model Merging.}
Model merging combines multiple models or task-specific adapters into a single model.
Representative approaches include Fisher-weighted merging, Model
Soups, and Task Arithmetic
~\cite{matena2022merging,wortsman2022model,ilharco2022editing}.
To mitigate task interference, TIES-Merging resolves
parameter-level conflicts, while DARE reduces adapter redundancy
through random dropping and rescaling
~\cite{yadav2023ties,yu2024language}.
Other studies investigate merge scaling and importance-aware
weighting~\cite{yadav2024matters,lee2025dynamic}.
Core Space merging aligns LoRA adapters in a shared low-rank basis
~\cite{panariello2026accurate}, whereas TSV uses task singular
directions to characterize and reduce cross-task interference
~\cite{gargiulo2025task}.

\begin{figure*}[htp!]
    \centering
    \includegraphics[width=0.90\textwidth]{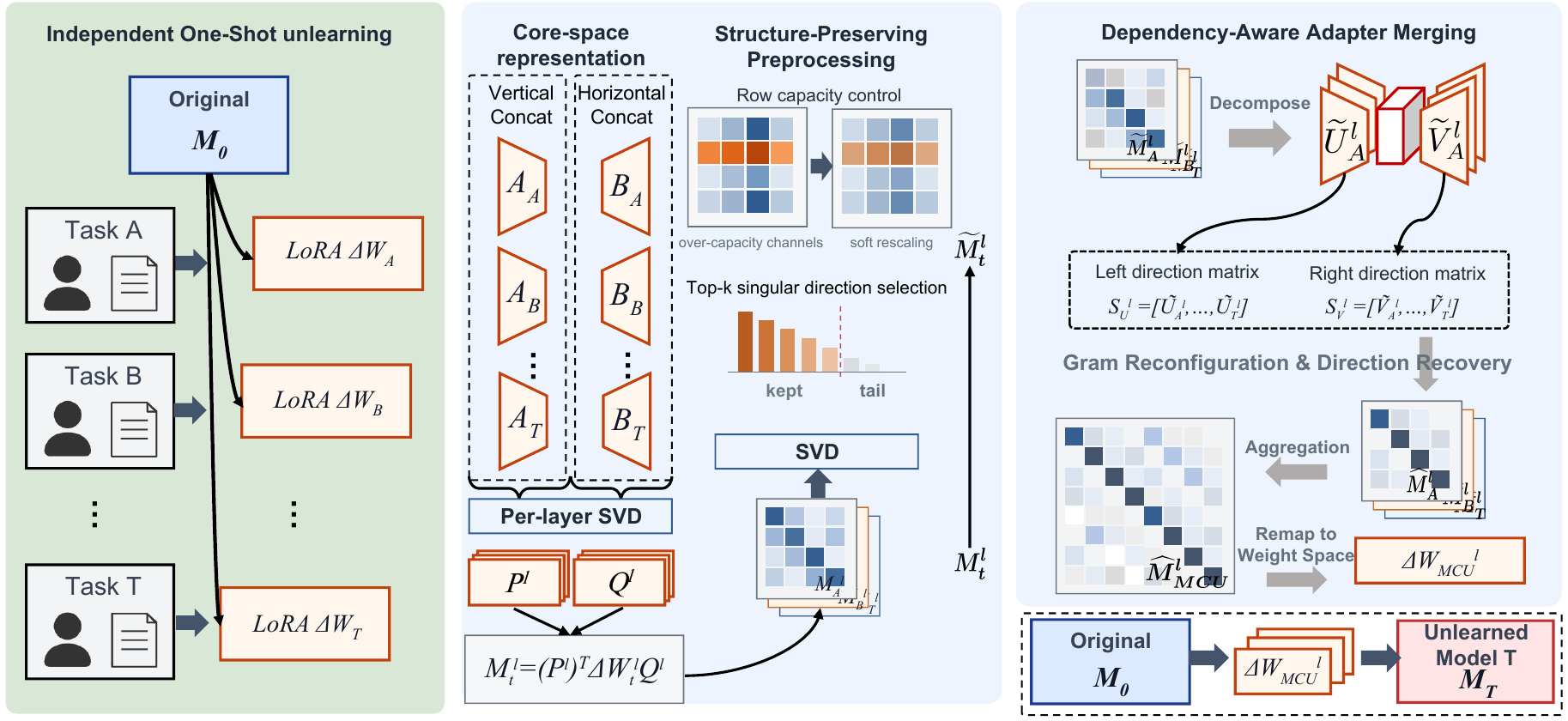}
    \vspace{-0.8em}
    \caption{Overall framework of MCU. (1) One-shot unlearning LoRA adapters derived from the same base model are projected into a shared space and decomposed into task singular directions. (2) Dominant direction selection and channel capacity control preserve the principal task structure while discarding low-contribution tail directions and rescaling over-concentrated core-space coordinates. (3) The processed task directions undergo dependency-aware Gram reconfiguration and Procrustes recovery before being merged in the core space and mapped back to the original parameter space.
    }
    \label{fig2}
    \vspace{-1.8em}
\end{figure*}

\vspace{-1.2em}

\section{Method}
\subsection{Adapter Merging and Cross-Task Dependencies}
\label{sec:continual_merging}
Following the merging-based perspective introduced above, we first formalize the continual adapter-merging setting and characterize the cross-task dependencies that motivate the design of MCU.

\vspace{-0.5em}
\paragraph{Dynamic adapter merging.}
Let $\mathcal{M}_0$ denote a pretrained MLLM with parameters
$\theta_0$, and let
$\mathcal{T}
=
\{(\mathcal{D}_f^t,\mathcal{D}_r^t)\}_{t=1}^{T}.$
denote a sequence of unlearning requests.
For each request $t$, an existing one-shot unlearning method
independently trains a LoRA adapter from the same base model
$\mathcal{M}_0$.
For the $l$-th linear layer, the resulting parameter update is
\begin{equation}
\Delta W_t^l
=
B_t^l A_t^l,
\quad
W_t^l=W_0^l+\Delta W_t^l,
\label{eq:one_shot_lora}
\end{equation}
where $A_t^l$ and $B_t^l$ are the low-rank LoRA factors, with the
standard LoRA scaling absorbed into the update~\cite{hu2022lora}.
Because all adapters are learned relative to the same initialization, their updates can be directly compared and merged.

At continual step $s$, after receiving the first $s$ requests, MCU
dynamically merges the corresponding one-shot adapters into a
unified adapter:
\begin{align}
\Delta W_{\mathrm{MCU},s}^l &= \operatorname{Merge}\left(\Delta W_1^l,\ldots,\Delta W_s^l\right), \\
W_{\star,s}^l &= W_0^l+\alpha\Delta W_{\mathrm{MCU},s}^l.
\label{eq:continual_adapter_merging}
\end{align}
where $\alpha$ is a global merging coefficient.
The objective is to preserve the unlearning effects of the available
one-shot adapters while maintaining retained knowledge and general
multimodal utility.

\vspace{-0.5em}
\paragraph{Shared core-space representation.}
Directly manipulating full parameter adapters is inefficient and
obscures their low-rank structures.
Following Core Space Merging~\cite{panariello2026accurate}, at step
$s$ we construct orthonormal bases $P_s^l$ and $Q_s^l$ spanning the
joint column and row spaces of
$\{\Delta W_t^l\}_{t=1}^{s}.$
Each adapter is represented in this shared coordinate system as
\begin{equation}
M_t^l
=
\left(P_s^l\right)^\top
\Delta W_t^l
Q_s^l,
\quad
\Delta W_t^l
=
P_s^l M_t^l
\left(Q_s^l\right)^\top.
\label{eq:core_space_representation}
\end{equation}
Since the bases span the joint update subspaces, the representation
is reversible up to numerical precision.
The core matrices $\{M_t^l\}_{t=1}^{s}$ therefore retain the original LoRA adapters in a compact and aligned space, where their spectral
structures and cross-task dependencies can be analyzed efficiently.
For readability, we omit the continual-step subscript of $P_s^l$
and $Q_s^l$ in the remainder of the paper.

\begin{figure}[t]
    \centering
    \includegraphics[width=0.95\linewidth]{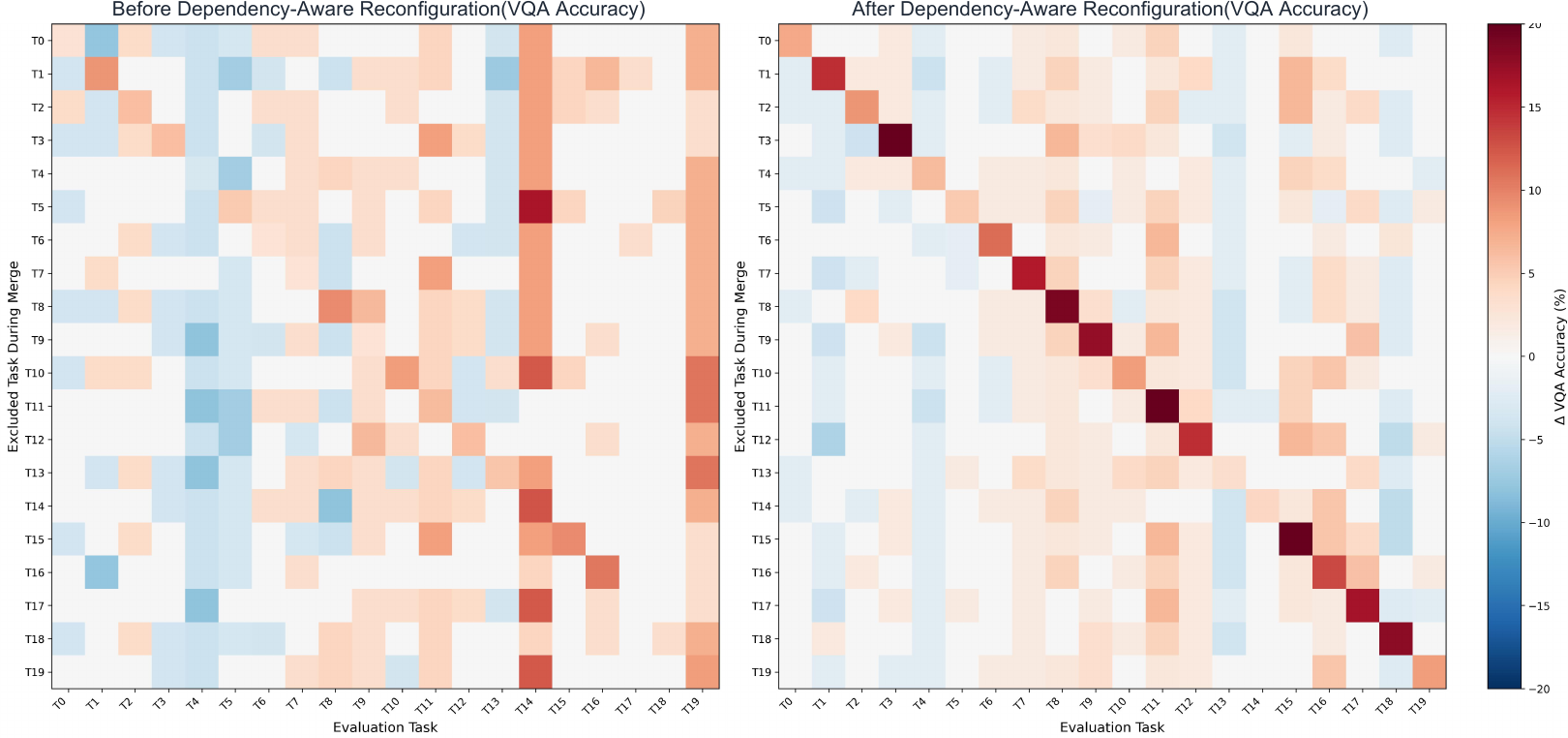}
    \vspace{-0.8em}
    \caption{Leave-one-task-out VQA analysis before and after dependency-aware reconfiguration. Columns indicate the excluded adapters, and rows indicate the evaluation targets. Each cell reports the accuracy change relative to the corresponding one-shot unlearned model.
}
    \label{fig3}
    \vspace{-1.8em}
\end{figure}

\label{sec:cross_task_dependency}
\vspace{-0.5em}
\paragraph{Cross-task unlearning transfer.}
Although the one-shot adapters are trained independently, their unlearning effects need not be isolated across requests.
To examine their interactions, we conduct a leave-one-out merging study: for each target request $t$, we exclude its corresponding adapter, merge the remaining adapters, and evaluate the resulting model on $\mathcal{D}_f^t$.
If the adapters encoded disjoint unlearning effects, excluding the target adapter would largely restore the base-model behavior on the held-out target.
Instead, the merged model still exhibits a non-negligible unlearning effect, as shown in Fig.~\ref{fig3}.
We refer to this phenomenon as \emph{cross-task unlearning transfer}.
It indicates that independently trained one-shot adapters exhibit cross-task dependencies whose effects can transfer across unlearning requests.

\vspace{-0.5em}
\paragraph{Task-direction dependencies.}
We characterize these shared structures through the singular
directions of the aligned core updates.
For request $t$ at layer $l$, we decompose
\begin{equation}
M_t^l
=
U_t^l\Sigma_t^l
\left(V_t^l\right)^\top,
\label{eq:task_svd}
\end{equation}
where $U_t^l$ and $V_t^l$ contain the left and right singular
directions, respectively.
Because all $M_t^l$ are expressed in the same core bases, these
directions are comparable across requests.
At continual step $s$, we concatenate them as
\begin{equation}
S_U^l
=
\left[U_1^l,\ldots,U_s^l\right],
\quad
S_V^l
=
\left[V_1^l,\ldots,V_s^l\right],
\label{eq:direction_concatenation}
\end{equation}
and define the corresponding Gram matrices
\begin{equation}
G_U^l
=
\left(S_U^l\right)^\top S_U^l,
\quad
G_V^l
=
\left(S_V^l\right)^\top S_V^l.
\label{eq:direction_gram}
\end{equation}
Their off-diagonal task blocks encode cross-request directional
dependencies.

Importantly, a cross-task dependency may be synergistic or interfering, and its magnitude alone does not distinguish between the two.
For component $a$ of request $i$ and component $b$ of request $j$,
their normalized matrix interaction is
\begin{equation}
\chi_{ij,ab}^l
=
\left[\left(U_i^l\right)^\top U_j^l\right]_{ab}
\left[\left(V_i^l\right)^\top V_j^l\right]_{ab}.
\label{eq:component_interaction}
\end{equation}
Indeed, the Frobenius inner product between the corresponding
rank-one update components equals
$\sigma_{i,a}^l\sigma_{j,b}^l\chi_{ij,ab}^l$.
Positive values indicate geometrically aligned rank-one updates,
whereas negative values indicate parameter-space interference that
may cause cancellation during merging.
Since directly constraining $\chi_{ij,ab}^l$ couples the left and
right geometries, MCU adopts a separable sufficient surrogate that
targets non-negative cross-request similarities on both sides while
remaining close to the original Gram geometry.
Unlike either singular-vector similarity alone,
$\chi_{ij,ab}^l$ is invariant to the paired sign ambiguity of the
singular value decomposition.


\vspace{-0.5em}
\subsection{Merging for Continual Unlearning}
\label{sec:mcu}

\paragraph{Overview.}
Figure~\ref{fig2} illustrates the overall procedure of MCU.
Given the one-shot unlearning adapters available at continual step
$s$, MCU first preserves the dominant structure of each core update,
then controls over-concentrated core-space coordinates, and finally
reconfigures their cross-request directional dependencies before
merging.
The first two operations produce structured updates that retain the
principal information of the individual adapters while reducing
unnecessary complexity for the subsequent joint reconfiguration.

\vspace{-0.5em}
\paragraph{Dominant direction selection.}
The singular spectra of unlearning adapters are typically concentrated,
with low-energy tail components contributing limited update magnitude
while increasing the number of directions involved in cross-request
reconfiguration.
MCU therefore retains only the leading $k$ singular components of
each core update.
Using the decomposition in Eq.~\eqref{eq:task_svd}, we define
\begin{equation}
\overline{M}_t^l
=
U_{t,k}^l
\Sigma_{t,k}^l
\left(V_{t,k}^l\right)^\top,
\label{eq:dominant_selection}
\end{equation}
where $U_{t,k}^l$ and $V_{t,k}^l$ contain the leading $k$ left and
right singular directions, respectively, and $\Sigma_{t,k}^l$
contains their associated singular values.
By the optimality of truncated SVD, $\overline{M}_t^l$ is the closest
rank-$k$ approximation to $M_t^l$ under the Frobenius norm, reducing
the number of low-contribution directions involved in joint
reconfiguration.

\vspace{-0.5em}
\paragraph{Channel capacity control.}
Dominant direction selection controls spectral complexity but does
not regulate how the retained update is distributed across
core-space coordinates.
In practice, the row norms of $\overline{M}_t^l$ exhibit a long-tailed
distribution, such that a small number of coordinates can dominate
the geometry of the update and its subsequent merging behavior.
MCU applies a soft row-capacity constraint to reduce this
over-concentration without removing the corresponding coordinates.

For row $i$ of $\overline{M}_t^l$, we define
\begin{equation}
\rho_{t,i}^l
=
\left\|
\overline{M}_t^l[i,:]
\right\|_2,
\quad
\tau_{t,q}^l
=
\operatorname{Quantile}_{q}
\left(
\left\{
\rho_{t,i}^l
\right\}_{i}
\right).
\label{eq:capacity_threshold}
\end{equation}
where $q\in(0,1)$ determines the capacity threshold.
The structured update is then obtained by
\begin{equation}
\widetilde{M}_t^l[i,:]
=
\min
\left(
1,
\frac{\tau_{t,q}^l}
{\rho_{t,i}^l+\epsilon}
\right)
\overline{M}_t^l[i,:],
\label{eq:row_capacity}
\end{equation}
where $\epsilon$ is a small constant for numerical stability.
Rows below the threshold remain unchanged, whereas rows above it are
rescaled to the threshold.
Dominant direction selection and channel capacity control therefore
operate on complementary structures: the former limits spectral
complexity, while the latter prevents a few core-space coordinates
from disproportionately dominating the retained update.

\vspace{-0.5em}
\paragraph{Dependency-aware direction reconfiguration.}
Although the structured updates preserve their dominant task-specific information, dependencies remain among their cross-request singular directions.
Rather than enforcing complete orthogonality, MCU suppresses antagonistic interactions while preserving the original geometry and beneficial dependencies that support cross-task unlearning transfer.
We first decompose each structured update as
\begin{equation}
\widetilde{M}_t^l
=
\widetilde{U}_t^l
\widetilde{\Sigma}_t^l
\left(\widetilde{V}_t^l\right)^\top .
\label{eq:structured_svd}
\end{equation}
After fixing the paired signs of the singular components using a
deterministic orientation rule solely for reproducibility, we
concatenate the left and right directions as
\begin{equation}
\widetilde{S}_U^l
=
\left[
\widetilde{U}_1^l,\ldots,\widetilde{U}_s^l
\right],
\qquad
\widetilde{S}_V^l
=
\left[
\widetilde{V}_1^l,\ldots,\widetilde{V}_s^l
\right].
\label{eq:structured_direction_concatenation}
\end{equation}
For $X\in\{U,V\}$, the corresponding Gram matrix is
\begin{equation}
G_{X,0}^l
=
\left(\widetilde{S}_X^l\right)^\top
\widetilde{S}_X^l .
\label{eq:original_direction_gram}
\end{equation}

Directly constraining the sign-invariant interaction in
Eq.~\eqref{eq:component_interaction} jointly couples the left and
right geometries.
We therefore adopt a tractable separable surrogate that requires
non-negative cross-request similarities on both sides.
This condition is sufficient, but not necessary, for a non-interfering rank-one interaction.
Let $\mathcal{I}_t^l$ denote the indices of the directions belonging
to request $t$ at layer $l$, and define the set of cross-request
direction pairs as
\begin{equation}
\mathcal{C}^l
=
\left\{
(p,q)
\;\middle|\;
p\in\mathcal{I}_i^l,\;
q\in\mathcal{I}_j^l,\;
i\neq j
\right\}.
\label{eq:cross_request_index_set}
\end{equation}
For $X\in\{U,V\}$, MCU solves
\begin{equation}
\setlength{\jot}{3pt}
\begin{aligned}
\widehat{G}_{X}^{l}=\arg\min_{G}& \left\|G-G_{X,0}^{l}\right\|_{F}^{2},\\
\mathrm{s.t.}\quad& G\succeq 0,\;\operatorname{rank}(G)\leq d_X^l,\; G_{pq}\geq 0,\;(p,q)\in\mathcal{C}^l,\\
& G\!\left[\mathcal{I}_t^l,\mathcal{I}_t^l\right]=I_{\lvert\mathcal{I}_t^l\rvert},\quad t=1,\ldots,s .
\end{aligned}
\label{eq:compatible_gram}
\end{equation}
Here, $d_U^l$ and $d_V^l$ denote the ambient dimensions of the left
and right direction spaces, respectively.
The proximity objective limits unnecessary changes introduced by the
conservative surrogate, while the identity-block constraints preserve
within-request orthonormality.

Because the rank-constrained feasible set is non-convex, we
approximately solve Eq.~\eqref{eq:compatible_gram} by alternating
projections onto the rank-constrained positive-semidefinite set, the
non-negative cross-request entries, and the within-request identity
blocks.
The sign invariance of the original interaction, the sufficient
nature and limitations of the separable surrogate, and the detailed
optimization procedure are provided in the supplementary material.

\definecolor{refgray}{RGB}{246,246,246}
\definecolor{MCUgreen}{RGB}{230,242,230}

\begin{table*}[ht]
\centering

\begingroup
\fontsize{7.4pt}{8.2pt}\selectfont
\setlength{\tabcolsep}{0.95pt}
\renewcommand{\arraystretch}{1.00}

\begin{tabular*}{\textwidth}{
@{\extracolsep{\fill}}
l
c
*{8}{c}
@{\hspace{4pt}}
*{8}{c}
@{}
}
\toprule

\multicolumn{1}{c}{\multirow{3}{*}{\textbf{Method}}}
& \multicolumn{1}{c}{\multirow{3}{*}{\textbf{Eval.}}}
& \multicolumn{8}{c}{\textbf{Qwen2-VL-7B}}
& \multicolumn{8}{c}{\textbf{LLaVA-1.5-7B}} \\

\cmidrule(lr){3-10}
\cmidrule(lr){11-18}

&
& \multicolumn{2}{c}{\textbf{Task 10}}
& \multicolumn{2}{c}{\textbf{Task 20}}
& \multicolumn{2}{c}{\textbf{Task 50}}
& \multicolumn{2}{c}{\textbf{Task 100}}
& \multicolumn{2}{c}{\textbf{Task 10}}
& \multicolumn{2}{c}{\textbf{Task 20}}
& \multicolumn{2}{c}{\textbf{Task 50}}
& \multicolumn{2}{c}{\textbf{Task 100}} \\

\cmidrule(lr){3-4}
\cmidrule(lr){5-6}
\cmidrule(lr){7-8}
\cmidrule(lr){9-10}
\cmidrule(lr){11-12}
\cmidrule(lr){13-14}
\cmidrule(lr){15-16}
\cmidrule(lr){17-18}

&
& \textbf{Forget}\,$\downarrow$ & \textbf{Retain}\,$\uparrow$
& \textbf{Forget}\,$\downarrow$ & \textbf{Retain}\,$\uparrow$
& \textbf{Forget}\,$\downarrow$ & \textbf{Retain}\,$\uparrow$
& \textbf{Forget}\,$\downarrow$ & \textbf{Retain}\,$\uparrow$
& \textbf{Forget}\,$\downarrow$ & \textbf{Retain}\,$\uparrow$
& \textbf{Forget}\,$\downarrow$ & \textbf{Retain}\,$\uparrow$
& \textbf{Forget}\,$\downarrow$ & \textbf{Retain}\,$\uparrow$
& \textbf{Forget}\,$\downarrow$ & \textbf{Retain}\,$\uparrow$ \\

\midrule

\rowcolor{refgray}
& VQA
& 88.1 & 88.9
& 87.9 & 88.4
& 89.0 & 88.6
& 89.1 & 88.1
& 66.8 & 66.2
& 64.4 & 67.1
& 67.4 & 68.0
& 65.8 & 67.2 \\

\rowcolor{refgray}
\multirow{-2}{*}{\textit{Vanilla}}
& QA
& 95.5 & 94.6
& 96.5 & 94.8
& 96.1 & 95.2
& 93.1 & 94.2
& 87.8 & 81.1
& 85.3 & 82.5
& 85.1 & 81.9
& 86.8 & 82.6 \\

\midrule

\multirow{2}{*}{GA}
& VQA
& \NA & \NA
& \NA & \NA
& \NA & \NA
& \NA & \NA
& 0.0 & 17.0
& 0.3 & 0.1
& \NA & \NA
& \NA & \NA \\

& QA
& \NA & \NA
& \NA & \NA
& \NA & \NA
& \NA & \NA
& 5.9 & 13.2
& \NA & \NA
& \NA & \NA
& \NA & \NA \\

\addlinespace[0.08em]

\multirow{2}{*}{GA-Diff}
& VQA
& 69.3 & 71.6
& 21.8 & 60.0
& 15.2 & 47.8
& 25.3 & 45.5
& 49.7 & 46.5
& 42.7 & 42.1
& 38.5 & 42.7
& 34.2 & 41.2 \\

& QA
& 78.1 & 86.9
& 52.8 & 84.1
& 45.8 & 46.3
& 66.7 & 67.5
& 75.6 & 71.5
& 70.0 & 69.4
& 65.2 & 68.4
& 65.2 & 66.8 \\

\addlinespace[0.08em]

\multirow{2}{*}{KL-Min}
& VQA
& 46.6 & 79.3
& 24.3 & 43.8
& \NA & 13.8
& \NA & \NA
& 58.8 & 63.3
& 63.2 & 60.9
& 57.4 & 59.0
& 46.6 & 53.6 \\

& QA
& 44.3 & 74.3
& 14.3 & 28.6
& 3.8 & 11.8
& \NA & \NA
& 77.3 & 78.7
& 81.3 & 80.8
& 78.1 & 77.4
& 73.7 & 71.5 \\

\multirow{2}{*}{NPO}
& VQA
& \NA & \NA
& \NA & \NA
& \NA & \NA
& \NA & \NA
& \NA & \NA
& \NA & \NA
& \NA & \NA
& \NA & \NA \\

& QA
& \NA & \NA
& \NA & \NA
& \NA & \NA
& \NA & \NA
& \NA & \NA
& \NA & \NA
& \NA & \NA
& \NA & \NA \\

\midrule

\multirow{2}{*}{MANU}
& VQA
& 71.9 & 71.5
& 60.0 & 59.8
& 23.9 & 25.5
& 21.8 & 25.2
& 56.0 & 60.3
& 54.7 & 57.1
& 24.4 & 24.1
& 24.2 & 23.8 \\

& QA
& 80.4 & 93.2
& 75.6 & 93.4
& 48.3 & 45.0
& 45.4 & 44.3
& 84.1 & 80.4
& 83.2 & 81.2
& 38.7 & 37.1
& 32.3 & 34.3 \\

\addlinespace[0.08em]

\multirow{2}{*}{MMU}
& VQA
& 45.8 & 67.6
& 48.5 & 66.1
& 39.0 & 46.6
& 33.3 & 42.7
& 38.9 & 48.2
& 40.6 & 51.4
& 35.7 & 54.0
& 32.5 & 57.3 \\

& QA
& 72.7 & 79.3
& 60.6 & 81.8
& 52.3 & 42.7
& 38.5 & 46.5
& 54.8 & 68.8
& 50.3 & 56.0
& 43.8 & 68.4
& 33.3 & 58.8 \\

\midrule










\rowcolor{MCUgreen}
& \textbf{VQA}
& 23.7 & 91.6
& 29.4 & 86.1
& 31.4 & 85.4
& 28.7 & 75.6
& 37.3 & 67.6
& 35.3 & 64.5
& 25.3 & 67.4
& 24.4 & 60.6 \\

\rowcolor{MCUgreen}
\multirow{-2}{*}{\textbf{MCU}}
& \textbf{QA}
& 34.5 & 80.3
& 33.4 & 73.0
& 36.5 & 70.6
& 30.9 & 75.2

& 41.3 & 71.0
& 39.7 & 68.4
& 23.9 & 67.8
& 23.7 & 66.8
\\

\bottomrule
\end{tabular*}

\endgroup
\vspace{-0.5em}
\caption{
Current-batch unlearning and retention results on ICU-Bench after 10, 20, 50, and 100 unlearning requests.
Each method is evaluated using VQA and QA accuracy.
Lower Forget and higher Retain indicate better performance.
\textbf{\NA} denotes unavailable or invalid results due to unstable
optimization or model collapse.
}
\label{tab1}
\vspace{-0.8em}
\end{table*}

\begin{figure*}[htp!]
    \centering
    \includegraphics[width=0.85\textwidth]{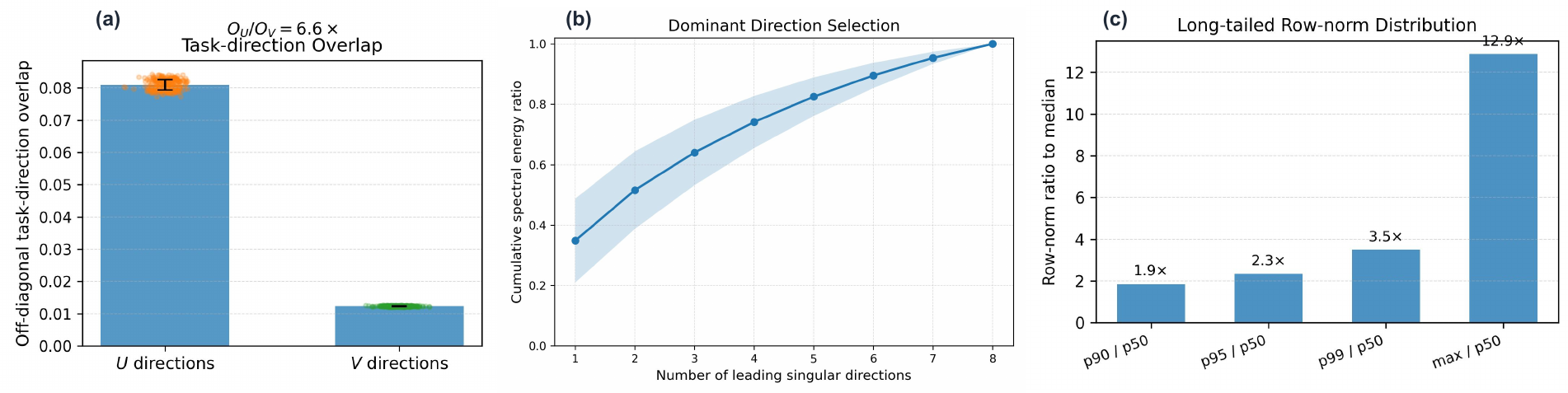}
    \vspace{-0.8em}
    \caption{
(a) Off-diagonal task-direction overlap of the singular subspaces.
unlearning adapters exhibit substantially stronger overlap on the $U$-side than on the $V$-side, with $O_U/O_V=6.6\times$, indicating stronger output-side cross-task dependencies.
(b) Cumulative spectral energy of one-shot unlearning adapters.
The curve reports the mean cumulative energy across all task--module updates, while the shaded region denotes one standard deviation.
The leading singular components capture most of the update energy: the top-4 directions retain $74.1\%$ energy on average, while the top-6 directions retain $89.6\%$.
(c) Row-norm concentration after spectral selection.
The retained updates show a long-tailed coordinate distribution, where the 99th-percentile row norm is $3.5\times$ the median and the maximum row norm reaches $12.9\times$ the median.
    }
    \label{fig4}
    \vspace{-0.8em}
\end{figure*}

\vspace{-0.5em}
\paragraph{Direction recovery and merging.}
The optimized Gram matrices specify the desired pairwise geometry but
do not directly provide direction matrices in the original core
space.
For $X\in\{U,V\}$, we factor
\begin{equation}
\widehat{G}_X^l
=
\left(R_X^l\right)^\top R_X^l
\label{eq:gram_factorization}
\end{equation}
and recover the closest realization to
$\widetilde{S}_X^l$ through an orthogonal Procrustes problem:
\begin{equation}
O_X^{l,\star}
=
\arg\min_{O^\top O=I}
\big\|
OR_X^l-\widetilde{S}_X^l
\big\|_F^2,
\quad
\widehat{S}_X^l
=
O_X^{l,\star}R_X^l.
\label{eq:procrustes_recovery}
\end{equation}
The recovered directions satisfy
$\left(\widehat{S}_X^l\right)^\top\widehat{S}_X^l
=\widehat{G}_X^l$
while remaining as close as possible to their structured
counterparts under the chosen factorization. Because the within-request Gram blocks are fixed to identity in Eq.~\eqref{eq:compatible_gram}, the recovered direction blocks remain orthonormal within each request.
We partition $\widehat{S}_U^l$ and $\widehat{S}_V^l$ according to
their original request blocks, obtaining $\{\widehat{U}_t^l\}_{t=1}^{s}$ and $\{\widehat{V}_t^l\}_{t=1}^{s}$.
Each processed core update is reconstructed using the structured singular-value coefficients:
\begin{equation}
\widehat{M}_t^l
=
\widehat{U}_t^l
\widetilde{\Sigma}_t^l
\left(\widehat{V}_t^l\right)^\top.
\label{eq:merging_compatible_update}
\end{equation}


\vspace{-0.4em}
Finally, the request-specific updates are merged in the shared core
space and mapped back to the original parameter space:
\begin{equation}
M_{\mathrm{MCU},s}^l
=
\sum_{t=1}^{s}
\widehat{M}_t^l,
\quad
\Delta W_{\mathrm{MCU},s}^l
=
P^l
M_{\mathrm{MCU},s}^l
\left(Q^l\right)^\top.
\label{eq:mcu_core_merging}
\end{equation}
Applying the procedure to all LoRA-enabled layers yields the unified
model $\mathcal{M}_{\theta_s^\star}$ at continual step $s$.
Given the one-shot adapters, MCU is a training-free, post-hoc merging
procedure and requires no additional gradient-based optimization.

\vspace{-0.8em}
\section{Experiments}

\begin{figure*}[ht]
    \centering
    \includegraphics[width=0.95\linewidth]{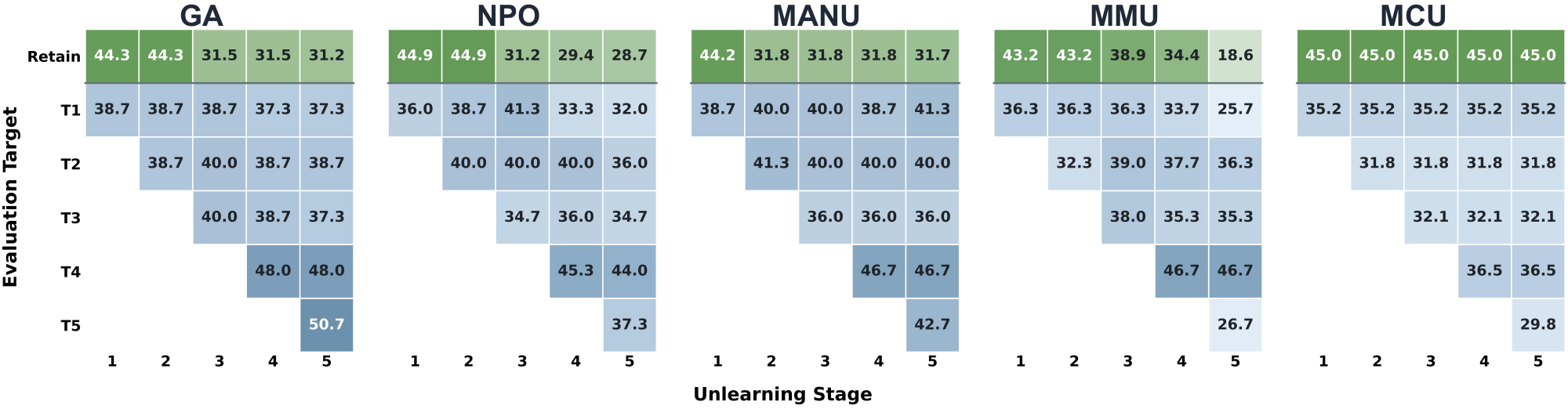}  
    \vspace{-0.8em}
    \caption{Stage-wise VQA performance under the five-stage continual unlearning protocol on MLLMU-Bench. Columns denote unlearning stages, while rows denote the fixed Retain set and task-specific unlearning targets. Lower target accuracy indicates stronger unlearning, whereas stable Retain accuracy reflects better utility preservation. Blank cells correspond to stages before a target is introduced.}    
    \label{fig5}
    \vspace{-1.8em}
\end{figure*}

\subsection{Experimental Setup}
\label{sec:experimental_setup}

\noindent\textbf{Benchmarks and protocols.}
We evaluate MCU on ICU-Bench~\cite{wang2026icu} and
MLLMU-Bench~\cite{liu2025protecting}.
For ICU-Bench, we follow the official 100-task protocol, with every
ten tasks forming one batch; Forget is evaluated after each task,
while Retain and in-domain utility are evaluated after each batch.
For MLLMU-Bench, we partition the 15\% Forget Set into five disjoint stages of 15 target profiles and evaluate the same retain subset throughout the sequence. The complete split and evaluation protocol are provided in the
supplementary material.

\begin{table}[tp!]
\centering
\scriptsize
\setlength{\tabcolsep}{1.2pt}
\renewcommand{\arraystretch}{0.88}
\resizebox{\linewidth}{!}{
\begin{tabular}{l|cccc|cccc}
\toprule
\multirow{2}{*}{\textbf{Method}}
& \multicolumn{4}{c|}{\textbf{Qwen2-VL-7B}}
& \multicolumn{4}{c}{\textbf{LLaVA-1.5-7B}} \\
\cmidrule(lr){2-5}\cmidrule(lr){6-9}
& \textbf{VQA-F} $\downarrow$
& \textbf{VQA-R} $\uparrow$
& \textbf{QA-F} $\downarrow$
& \textbf{QA-R} $\uparrow$
& \textbf{VQA-F} $\downarrow$
& \textbf{VQA-R} $\uparrow$
& \textbf{QA-F} $\downarrow$
& \textbf{QA-R} $\uparrow$ \\
\midrule

Task Arithmetic
& 42.8 & 76.5 & 47.9 & 61.3
& 45.6 & 35.2 & 55.8 & 42.1 \\

TIES
& 39.5 & 78.8 & 44.2 & 64.7
& 43.8 & 36.9 & 53.4 & 43.6 \\

DARE
& 36.7 & 80.1 & 41.8 & 66.5
& 42.2 & 37.5 & 52.1 & 44.0 \\

TSV
& 34.9 & \underline{81.7} & 39.9 & \underline{67.2}
& 40.8 & \underline{38.2} & 50.3 & \underline{45.1} \\

Core-space
& 34.1 & 72.5 & 39.6 & 58.6
& 35.6 & 35.2 & 44.8 & 44.7 \\
\midrule

\rowcolor{lightblue}
\textbf{MCU}
&  31.4  &  85.4 
&  36.5  &  70.6 
&  25.3  &  67.4 
&  23.9  &  67.8  \\

\bottomrule
\end{tabular}
}
\vspace{-0.5em}
\caption{
Comparison with representative merging methods on ICU-Bench after 50 unlearning tasks.
F and R denote Forget and Retain accuracy, respectively.
}
\label{tab:merging_comparison}
\vspace{-2.8em}
\end{table}

\noindent\textbf{Models and baselines.}
We evaluate LLaVA-1.5-7B~\cite{liu2024improved} and
Qwen2-VL-7B~\cite{wang2024qwen2}.
MCU independently trains one-shot GA-Diff LoRA adapters from the same
base checkpoint~\cite{liu2022continual}.
Sequential baselines include GA~\cite{thudi2022unrolling},
GA-Diff~\cite{liu2022continual}, KL-Min~\cite{maini2024tofu},
NPO~\cite{zhang2024negative}, MANU~\cite{liu2025modality}, and
MMUnlearner(MMU)~\cite{huo2025mmunlearner}.
Merging baselines include Task Arithmetic~\cite{ilharco2022editing},
TIES-Merging~\cite{yadav2023ties}, DARE~\cite{yu2024language},
TSV~\cite{gargiulo2025task}, and Core Space
merging~\cite{panariello2026accurate}; all use the same adapter bank.

\noindent\textbf{Metrics.}
We report VQA and QA accuracy on both benchmarks.
For ICU-Bench, we additionally report Current/Historical
Forget and Retain, in-domain utility, Generation Quality (GQ),
Retain Stability Rate (RSR), and Forgetting Rebound (FR).
For MLLMU-Bench, we report Current/Historical Forget and Retain performance.
Lower Forget, RSR, and FR are preferred, while higher values are
better for the remaining metrics.
Full configurations are provided in the supplementary material.

\subsection{Main Results}
\label{sec:main_results}

\paragraph{Continual unlearning on ICU-Bench.}
Table~\ref{tab1} reports Current Forget and Current Retain
performance at representative checkpoints of the 100-task
ICU-Bench sequence.
Existing methods exhibit an increasingly unstable
unlearning--retention trade-off as the sequence grows.
Gradient-based methods often achieve low Forget accuracy at the cost
of severe utility degradation, whereas preference-based and
multimodal-specific methods struggle to maintain consistent
unlearning and retention over long sequences.

MCU consistently establishes a stronger balance across both
backbones and all sequence lengths.
At Task 100 on Qwen2-VL-7B, MCU achieves Forget VQA/QA scores of
28.7/30.9 while preserving Retain VQA/QA scores of 75.6/75.2.
Compared with MMUnlearner, this reduces Forget accuracy by 4.6/7.6
points and improves Retain accuracy by 32.9/28.7 points.
On LLaVA-1.5-7B, MCU obtains Forget scores of 24.4/23.7 together
with Retain scores of 60.6/66.8, outperforming MMUnlearner by
8.1/9.6 points on Forget and 3.3/8.0 points on Retain.
The consistent advantage from short to long sequences shows that MCU
remains effective as the number of accumulated one-shot adapters
increases.

Importantly, low Forget accuracy does not always correspond to
successful unlearning.
MANU obtains competitive Forget scores at later checkpoints, but its
retained performance and generation quality collapse, showing that
its apparent unlearning efficacy is largely caused by broad model
degradation.
MCU instead maintains effective target removal together with
substantially stronger retained performance.

\begin{table}[tp!]
\centering
\scriptsize
\setlength{\tabcolsep}{1.2pt}
\renewcommand{\arraystretch}{0.8}
\resizebox{\linewidth}{!}{
\begin{tabular}{l|cccc|cccc}
\toprule
\multirow{2}{*}{\textbf{Method}}
& \multicolumn{4}{c|}{\textbf{50 Tasks}}
& \multicolumn{4}{c}{\textbf{100 Tasks}} \\
\cmidrule(lr){2-5}\cmidrule(lr){6-9}
& \textbf{RSR} $\downarrow$
& \textbf{FR} $\downarrow$
& \textbf{GQ-F} $\uparrow$
& \textbf{GQ-R} $\uparrow$
& \textbf{RSR} $\downarrow$
& \textbf{FR} $\downarrow$
& \textbf{GQ-F} $\uparrow$
& \textbf{GQ-R} $\uparrow$ \\
\midrule

GA
& -- & -- & -- & --
& -- & -- & -- & -- \\

GA-Diff
& 9.23 & 0.51 & 1.340 & 1.390
& 6.46 & 0.28 & 0.900 & 1.240 \\

NPO
& -- & -- & -- & --
& -- & -- & -- & -- \\


KL-Min
& 4.72 & 0.87 & 0.450 & 1.910
& 4.20 & 2.61 & 0.370 & 1.840 \\

MANU
& 11.42 & -- & 0.776 & 0.771
& 5.12 & 0.34 & 0.447 & 0.448 \\

MMU
& 2.62 & 1.95 & 1.832 & 1.827
& 6.82 & 5.45 & 1.804 & 1.826 \\
\midrule

\rowcolor{lightblue}
\textbf{MCU}
& 1.02 & 1.86 & 1.989 & 1.994
& 1.10 & 1.12 & 1.990 & 1.993\\
\bottomrule
\end{tabular}
}
\vspace{-0.5em}
\caption{
Sequence-level evaluation on Qwen2-VL-7B after 50 and 100
unlearning requests.
}
\label{tab2}
\vspace{-2.5em}
\end{table}

\paragraph{Continual unlearning on MLLMU-Bench.}
Figure~\ref{fig5} evaluates all methods under the five-stage continual protocol constructed from MLLMU-Bench. Because the Retain Set remains fixed, its stage-wise accuracy directly reflects retention drift.
Existing methods either suppress new targets by progressively degrading the retained sets or fail to preserve unlearning on historical targets.
In particular, MMUnlearner and NPO undergo substantial retention degradation over the sequence, whereas GA and MANU exhibit unstable responses on historical targets.
MCU instead maintains stable Retain performance while consistently suppressing both current and previously introduced target knowledge.
Together with the ICU-Bench results, this demonstrates the effectiveness of MCU across different sequence lengths and target knowledge types.

\begin{table}[t]
\centering
\scriptsize
\setlength{\tabcolsep}{1.9pt}
\renewcommand{\arraystretch}{0.88}

\begin{tabular}{l|cccc|cc|cc}
\toprule
\multirow{2}{*}{\textbf{Variant}}
& \multirow{2}{*}{\textbf{Sel.}}
& \multirow{2}{*}{\textbf{Cap.}}
& \multirow{2}{*}{\textbf{Orth.}}
& \multirow{2}{*}{\textbf{Reconf.}}
& \multicolumn{2}{c|}{\textbf{Forget} $\downarrow$}
& \multicolumn{2}{c}{\textbf{Retain} $\uparrow$} \\
\cmidrule(lr){6-7}
\cmidrule(lr){8-9}
& & & &
& \textbf{VQA} & \textbf{QA}
& \textbf{VQA} & \textbf{QA} \\
\midrule

Direct Addition
&  &  &  &
& 47.9 & 51.6
& 80.7 & 70.5 \\

Reconfiguration Only
&  &  &  & \checkmark
& 43.5 & 46.7
& 78.6 & 68.3 \\

Selection + Reconf.
& \checkmark &  &  & \checkmark
& 33.5 & 38.9
& 83.1 & 68.9 \\

Capacity + Reconf.
&  & \checkmark &  & \checkmark
& 34.3 & 39.8
& 82.2 & 68.1 \\

Shaping Only
& \checkmark & \checkmark &  &
& 36.0 & 41.2
& 84.2 & 69.8 \\

Shaping + Orth.
& \checkmark & \checkmark & \checkmark &
& 32.8 & 37.9
& 83.1 & 69.1 \\

\rowcolor{MCUgreen}
\textbf{MCU}
& \checkmark & \checkmark &  & \checkmark
& \textbf{31.4} & \textbf{36.5}
& \textbf{85.4} & \textbf{70.6} \\

\bottomrule
\end{tabular}

\caption{Component and direction-handling ablation on ICU-Bench with Qwen2-VL-7B after 50 unlearning tasks. Sel., Cap., Orth., and Reconf. denote direction selection, capacity control, strict orthogonalization, and dependency-aware reconfiguration, respectively.}
\label{tab:ablation_component}
\vspace{-2.8em}
\end{table}

\paragraph{Sequence-level stability.}
While Table~\ref{tab1} evaluates individual checkpoints,
Table~\ref{tab2} summarizes stability over the complete ICU-Bench
sequence.
At Task 100, MCU obtains an FR of 1.12 and an RSR of 1.10,
reducing historical rebound and retention drift by 4.33 and 5.72
points relative to MMUnlearner, respectively.
MCU also maintains GQ-F/GQ-R scores of 1.990/1.993, confirming
that its low Forget accuracy is not caused by invalid or collapsed
generation.
Thus, MCU preserves historical unlearning, retained performance, and
normal response quality throughout long request sequences.

\subsection{Ablation and Analysis}
\label{sec:ablation_analysis}

\paragraph{Cross-task unlearning transfer and update geometry.}
Figure~\ref{fig3} reports the leave-one-task-out analysis in the 20-task setting. For each column, we remove the adapter of the corresponding task, merge the remaining adapters, and evaluate the resulting model on all targets shown by the rows.
Each cell reports the VQA accuracy change relative to the corresponding one-shot unlearned model.
Before direction reconfiguration, removing a target adapter often produces only limited recovery on the diagonal, indicating that the remaining adapters still transfer non-negligible unlearning effects to the held-out target.
Complete results and additional analysis are provided in the
supplementary material.
Figure~\ref{fig4}(a) provides a geometric explanation: their singular
directions exhibit non-negligible cross-task dependencies, with
$\mathcal{O}_{U}=0.081$ and $\mathcal{O}_{V}=0.012$.
After dependency-aware direction reconfiguration, most tasks
exhibit stronger unlearning performance, while removing a target
adapter produces a clearer recovery on its corresponding target.
This indicates that the reconfigured geometry reduces harmful
cross-task interference while limiting disruption to synergistic dependencies.

Figure~\ref{fig4}(b) further shows that the leading four and six
singular components preserve 74.1\% and 89.6\% of the update energy,
respectively, supporting the removal of low-contribution tail
directions.
Figure~\ref{fig4}(c) reveals a strongly long-tailed row-norm
distribution: the 99th-percentile and maximum row norms are
3.5$\times$ and 12.9$\times$ the median, respectively.
These results motivate dominant direction selection and channel
capacity control before joint reconfiguration.
Complete leave-one-out results and geometric statistics are reported
in supplementary material.

\vspace{-0.5em}
\paragraph{Comparison with model merging methods.}
As shown in Table~\ref{tab:merging_comparison}, MCU achieves stronger unlearning while preserving more retained knowledge across both backbones.
By reconfiguring cross-task singular directions in the shared core space, MCU reduces cross-task interference that weakens unlearning while limiting disruption to synergistic dependencies across requests.
Conventional merging methods instead treat the parameter updates
induced by unlearning adapters as ordinary task vectors and fail to account for their distinct suppressive geometry.

\vspace{-0.5em}
\paragraph{Component ablation.}
Table~\ref{tab:ablation_component} evaluates the contribution of
update shaping and cross-task direction handling after 50 unlearning
tasks.
Compared with Direct Addition, dependency-aware reconfiguration lowers Forget VQA/QA from 47.9/51.6 to 43.5/46.7,
but also reduces Retain from 80.7/70.5 to 78.6/68.3.
Thus, directly reconfiguring the complete direction space strengthens
unlearning but can disrupt the original update structures.

Adding direction selection substantially improves both sides of the
trade-off, achieving Forget VQA/QA scores of 33.5/38.9 and Retain
scores of 83.1/68.9.
Capacity control similarly improves Forget to 34.3/39.8 while
recovering Retain VQA to 82.2.
These results show that removing low-contribution directions and
limiting over-concentrated coordinates provide more suitable updates
for subsequent joint reconfiguration.

The remaining variants clarify the role of cross-task direction
handling.
Shaping Only preserves strong Retain performance but leaves
cross-task interactions insufficiently resolved, resulting in weaker
Forget scores of 36.0/41.2.
Strict orthogonalization improves unlearning, but indiscriminately removes both interfering and synergistic dependencies.
Under the same shaping operations, MCU improves over strict
orthogonalization by 1.4/1.4 points on Forget and 2.3/1.5 points on
Retain.
The complete MCU therefore achieves the best Forget VQA/QA scores of
31.4/36.5 and the best Retain scores of 85.4/70.6, confirming the complementary roles of direction selection, capacity control, and dependency-aware reconfiguration.

\section{Conclusion}

In this work, we introduced MCU, a model-merging framework for continual multimodal unlearning that consolidates accumulated one-shot unlearning adapters into a unified model update.
Our analysis reveals cross-task dependencies that can either support beneficial unlearning transfer or induce antagonistic interactions, motivating MCU to preserve useful shared structures while suppressing harmful interference. 
Extensive experiments demonstrate that MCU achieves effective current and historical unlearning while preserving retained knowledge and general multimodal utility. Given the one-shot adapters, MCU requires no additional gradient-based optimization during merging and directly produces a unified update for deployment.

\newpage

\bibliography{aaai2027}

@inproceedings{pi2024mllm,
  title={Mllm-protector: Ensuring mllm’s safety without hurting performance},
  author={Pi, Renjie and Han, Tianyang and Zhang, Jianshu and Xie, Yueqi and Pan, Rui and Lian, Qing and Dong, Hanze and Zhang, Jipeng and Zhang, Tong},
  booktitle={Proceedings of the 2024 Conference on Empirical Methods in Natural Language Processing},
  pages={16012--16027},
  year={2024}
}

@article{li2024digger,
  title={Digger: Detecting copyright content mis-usage in large language model training},
  author={Li, Haodong and Deng, Gelei and Liu, Yi and Wang, Kailong and Li, Yuekang and Zhang, Tianwei and Liu, Yang and Xu, Guoai and Xu, Guosheng and Wang, Haoyu},
  journal={arXiv preprint arXiv:2401.00676},
  year={2024}
}

@inproceedings{cohen2025performance,
  title={Performance gap in entity knowledge extraction across modalities in vision language models},
  author={Cohen, Ido and Gottesman, Daniela and Geva, Mor and Giryes, Raja},
  booktitle={Proceedings of the 63rd Annual Meeting of the Association for Computational Linguistics (Volume 1: Long Papers)},
  pages={29095--29108},
  year={2025}
}

@inproceedings{garg2020formalizing,
  title={Formalizing data deletion in the context of the right to be forgotten},
  author={Garg, Sanjam and Goldwasser, Shafi and Vasudevan, Prashant Nalini},
  booktitle={Annual International Conference on the Theory and Applications of Cryptographic Techniques},
  pages={373--402},
  year={2020},
  organization={Springer}
}

@article{gupta2021adaptive,
  title={Adaptive machine unlearning},
  author={Gupta, Varun and Jung, Christopher and Neel, Seth and Roth, Aaron and Sharifi-Malvajerdi, Saeed and Waites, Chris},
  journal={Advances in Neural Information Processing Systems},
  volume={34},
  pages={16319--16330},
  year={2021}
}

@article{sekhari2021remember,
  title={Remember what you want to forget: Algorithms for machine unlearning},
  author={Sekhari, Ayush and Acharya, Jayadev and Kamath, Gautam and Suresh, Ananda Theertha},
  journal={Advances in Neural Information Processing Systems},
  volume={34},
  pages={18075--18086},
  year={2021}
}

@article{li2026mlubench,
  title={MLUBench: A Benchmark for Lifelong Unlearning Evaluation in MLLMs},
  author={Li, He and Chi, Haoang and Wang, Qizhou and Mao, Yunxin and Zhang, Zhiheng and Tan, Jie and Liu, Tongliang and Yang, Wenjing and Han, Bo},
  journal={arXiv preprint arXiv:2606.12809},
  year={2026}
}

@article{wang2026icu,
  title={ICU-Bench: Benchmarking Continual Unlearning in Multimodal Large Language Models},
  author={Wang, Yuhang and Mei, Wenjie and Zhang, Junkai and He, Guangyu and Niu, Zhenxing and Gao, Haichang},
  journal={arXiv preprint arXiv:2605.05938},
  year={2026}
}

@inproceedings{gao2025large,
  title={On large language model continual unlearning},
  author={Gao, Chongyang and Wang, Lixu and Ding, Kaize and Weng, Chenkai and Wang, Xiao and Zhu, Qi},
  booktitle={International Conference on Learning Representations},
  volume={2025},
  pages={101772--101801},
  year={2025}
}

@inproceedings{shi2025muse,
  title={Muse: Machine unlearning six-way evaluation for language models},
  author={Shi, Weijia and Lee, Jaechan and Huang, Yangsibo and Malladi, Sadhika and Zhao, Jieyu and Holtzman, Ari and Liu, Daogao and Zettlemoyer, Luke and Smith, Noah and Zhang, Chiyuan},
  booktitle={International Conference on Learning Representations},
  volume={2025},
  pages={27797--27818},
  year={2025}
}

@article{kawakami2025pulse,
  title={Pulse: Practical evaluation scenarios for large multimodal model unlearning},
  author={Kawakami, Tatsuki and Egashira, Kazuki and Miyai, Atsuyuki and Irie, Go and Aizawa, Kiyoharu},
  journal={arXiv preprint arXiv:2507.01271},
  year={2025}
}

@inproceedings{liu2025protecting,
  title={Protecting privacy in multimodal large language models with mllmu-bench},
  author={Liu, Zheyuan and Dou, Guangyao and Jia, Mengzhao and Tan, Zhaoxuan and Zeng, Qingkai and Yuan, Yongle and Jiang, Meng},
  booktitle={Proceedings of the 2025 Conference of the Nations of the Americas Chapter of the Association for Computational Linguistics: Human Language Technologies (Volume 1: Long Papers)},
  pages={4105--4135},
  year={2025}
}

@inproceedings{thudi2022unrolling,
  title={Unrolling sgd: Understanding factors influencing machine unlearning},
  author={Thudi, Anvith and Deza, Gabriel and Chandrasekaran, Varun and Papernot, Nicolas},
  booktitle={2022 IEEE 7th European Symposium on Security and Privacy (EuroS\&P)},
  pages={303--319},
  year={2022},
  organization={IEEE}
}

@inproceedings{liu2022continual,
  title={Continual learning and private unlearning},
  author={Liu, Bo and Liu, Qiang and Stone, Peter},
  booktitle={Conference on Lifelong Learning Agents},
  pages={243--254},
  year={2022},
  organization={PMLR}
}

@article{maini2024tofu,
  title={Tofu: A task of fictitious unlearning for llms},
  author={Maini, Pratyush and Feng, Zhili and Schwarzschild, Avi and Lipton, Zachary C and Kolter, J Zico},
  journal={arXiv preprint arXiv:2401.06121},
  year={2024}
}

@article{rafailov2023direct,
  title={Direct preference optimization: Your language model is secretly a reward model},
  author={Rafailov, Rafael and Sharma, Archit and Mitchell, Eric and Manning, Christopher D and Ermon, Stefano and Finn, Chelsea},
  journal={Advances in neural information processing systems},
  volume={36},
  pages={53728--53741},
  year={2023}
}

@article{zhang2024negative,
  title={Negative preference optimization: From catastrophic collapse to effective unlearning},
  author={Zhang, Ruiqi and Lin, Licong and Bai, Yu and Mei, Song},
  journal={arXiv preprint arXiv:2404.05868},
  year={2024}
}

@inproceedings{huo2025mmunlearner,
  title={Mmunlearner: Reformulating multimodal machine unlearning in the era of multimodal large language models},
  author={Huo, Jiahao and Yan, Yibo and Zheng, Xu and Lyu, Yuanhuiyi and Zou, Xin and Wei, Zhihua and Hu, Xuming},
  booktitle={Findings of the Association for Computational Linguistics: ACL 2025},
  pages={7190--7206},
  year={2025}
}

@inproceedings{liu2025modality,
  title={Modality-aware neuron pruning for unlearning in multimodal large language models},
  author={Liu, Zheyuan and Dou, Guangyao and Yuan, Xiangchi and Zhang, Chunhui and Tan, Zhaoxuan and Jiang, Meng},
  booktitle={Proceedings of the 63rd Annual Meeting of the Association for Computational Linguistics (Volume 1: Long Papers)},
  pages={5913--5933},
  year={2025}
}

@article{li2024single,
  title={Single image unlearning: Efficient machine unlearning in multimodal large language models},
  author={Li, Jiaqi and Wei, Qianshan and Zhang, Chuanyi and Qi, Guilin and Du, Miaozeng and Chen, Yongrui and Bi, Sheng and Liu, Fan},
  journal={Advances in Neural Information Processing Systems},
  volume={37},
  pages={35414--35453},
  year={2024}
}

@article{wang2025mllm,
  title={MLLM Machine Unlearning via Visual Knowledge Distillation},
  author={Wang, Yuhang and Niu, Zhenxing and Ji, Haoxuan and He, Guangyu and Gao, Haichang and Hua, Gang},
  journal={arXiv preprint arXiv:2512.11325},
  year={2025}
}

@article{wang2026null,
  title={Null Space Constrained Contrastive Visual Forgetting for MLLM Unlearning},
  author={Wang, Yuhang and Niu, Zhenxing and Ji, Haoxuan and He, Guangyu and Zhang, Linlin and Gao, Haichang},
  journal={arXiv preprint arXiv:2605.05909},
  year={2026}
}

@article{cheng2024mu,
  title={Mu-bench: A multitask multimodal benchmark for machine unlearning},
  author={Cheng, Jiali and Amiri, Hadi},
  journal={arXiv preprint arXiv:2406.14796},
  year={2024}
}

@article{xu2025pebench,
  title={Pebench: A fictitious dataset to benchmark machine unlearning for multimodal large language models},
  author={Xu, Zhaopan and Zhou, Pengfei and Tang, Weidong and Ai, Jiaxin and Zhao, Wangbo and Wang, Kai and Peng, Xiaojiang and Shao, Wenqi and Yao, Hongxun and Zhang, Kaipeng},
  journal={arXiv preprint arXiv:2503.12545},
  year={2025}
}

@inproceedings{dontsov2025clear,
  title={Clear: Character unlearning in textual and visual modalities},
  author={Dontsov, Alexey and Korzh, Dmitrii and Zhavoronkin, Alexey and Mikheev, Boris and Bobkov, Denis and Alanov, Aibek and Rogov, Oleg and Oseledets, Ivan and Tutubalina, Elena},
  booktitle={Findings of the Association for Computational Linguistics: ACL 2025},
  pages={20582--20603},
  year={2025}
}

@article{wang2026umu,
  title={Umu-bench: Closing the modality gap in multimodal unlearning evaluation},
  author={Wang, Chengye and Li, Yuyuan and Feng, XiaoHua and Chen, Chaochao and Zheng, Xiaolin and Yin, Jianwei},
  journal={Advances in Neural Information Processing Systems},
  volume={38},
  year={2026}
}

@article{yu2025forgetme,
  title={Forgetme: Benchmarking the selective forgetting capabilities of generative models},
  author={Yu, Zhenyu and Idris, Mohd Yamani Idna and Wang, Pei and Xia, Yuelong and Xiang, Yong},
  journal={Engineering Applications of Artificial Intelligence},
  volume={161},
  pages={112087},
  year={2025},
  publisher={Elsevier}
}

@article{matena2022merging,
  title={Merging models with fisher-weighted averaging},
  author={Matena, Michael S and Raffel, Colin},
  journal={Advances in Neural Information Processing Systems},
  volume={35},
  pages={17703--17716},
  year={2022}
}

@inproceedings{wortsman2022model,
  title={Model soups: averaging weights of multiple fine-tuned models improves accuracy without increasing inference time},
  author={Wortsman, Mitchell and Ilharco, Gabriel and Gadre, Samir Ya and Roelofs, Rebecca and Gontijo-Lopes, Raphael and Morcos, Ari S and Namkoong, Hongseok and Farhadi, Ali and Carmon, Yair and Kornblith, Simon and others},
  booktitle={International conference on machine learning},
  pages={23965--23998},
  year={2022},
  organization={PMLR}
}

@article{ilharco2022editing,
  title={Editing models with task arithmetic},
  author={Ilharco, Gabriel and Ribeiro, Marco Tulio and Wortsman, Mitchell and Gururangan, Suchin and Schmidt, Ludwig and Hajishirzi, Hannaneh and Farhadi, Ali},
  journal={arXiv preprint arXiv:2212.04089},
  year={2022}
}

@article{yadav2023ties,
  title={Ties-merging: Resolving interference when merging models},
  author={Yadav, Prateek and Tam, Derek and Choshen, Leshem and Raffel, Colin A and Bansal, Mohit},
  journal={Advances in neural information processing systems},
  volume={36},
  pages={7093--7115},
  year={2023}
}

@inproceedings{yu2024language,
  title={Language models are super mario: Absorbing abilities from homologous models as a free lunch},
  author={Yu, Le and Yu, Bowen and Yu, Haiyang and Huang, Fei and Li, Yongbin},
  booktitle={Forty-first International Conference on Machine Learning},
  year={2024}
}

@article{yadav2024matters,
  title={What matters for model merging at scale?},
  author={Yadav, Prateek and Vu, Tu and Lai, Jonathan and Chronopoulou, Alexandra and Faruqui, Manaal and Bansal, Mohit and Munkhdalai, Tsendsuren},
  journal={arXiv preprint arXiv:2410.03617},
  year={2024}
}

@inproceedings{lee2025dynamic,
  title={Dynamic fisher-weighted model merging via bayesian optimization},
  author={Lee, Sanwoo and Liu, Jiahao and Wang, Qifan and Wang, Jingang and Cai, Xunliang and Wu, Yunfang},
  booktitle={Proceedings of the 2025 Conference of the Nations of the Americas Chapter of the Association for Computational Linguistics: Human Language Technologies (Volume 1: Long Papers)},
  pages={4923--4935},
  year={2025}
}

@article{panariello2026accurate,
  title={Accurate and efficient low-rank model merging in core space},
  author={Panariello, Aniello and Marczak, Daniel and Magistri, Simone and Porrello, Angelo and Twardowski, Bart{\l}omiej and Bagdanov, Andrew and Calderara, Simone and van de Weijer, Joost},
  journal={Advances in Neural Information Processing Systems},
  volume={38},
  pages={61793--61825},
  year={2026}
}

@inproceedings{gargiulo2025task,
  title={Task singular vectors: Reducing task interference in model merging},
  author={Gargiulo, Antonio Andrea and Crisostomi, Donato and Bucarelli, Maria Sofia and Scardapane, Simone and Silvestri, Fabrizio and Rodola, Emanuele},
  booktitle={Proceedings of the Computer Vision and Pattern Recognition Conference},
  pages={18695--18705},
  year={2025}
}

@article{hu2022lora,
  title={Lora: Low-rank adaptation of large language models.},
  author={Hu, Edward J and Shen, Yelong and Wallis, Phillip and Allen-Zhu, Zeyuan and Li, Yuanzhi and Wang, Shean and Wang, Liang and Chen, Weizhu and others},
  journal={Iclr},
  volume={1},
  number={2},
  pages={3},
  year={2022}
}

@inproceedings{liu2024improved,
  title={Improved baselines with visual instruction tuning},
  author={Liu, Haotian and Li, Chunyuan and Li, Yuheng and Lee, Yong Jae},
  booktitle={Proceedings of the IEEE/CVF conference on computer vision and pattern recognition},
  pages={26296--26306},
  year={2024}
}

@article{wang2024qwen2,
  title={Qwen2-vl: Enhancing vision-language model's perception of the world at any resolution},
  author={Wang, Peng and Bai, Shuai and Tan, Sinan and Wang, Shijie and Fan, Zhihao and Bai, Jinze and Chen, Keqin and Liu, Xuejing and Wang, Jialin and Ge, Wenbin and others},
  journal={arXiv preprint arXiv:2409.12191},
  year={2024}
}

@article{lin2026cata,
  title={CATA: Continual Machine Unlearning via Conflict-Averse Task Arithmetic},
  author={Lin, Shen and Dong, Junhao and Chen, Rongjie and Zhang, Xiaoyu and Xu, Li and Chen, Xiaofeng},
  journal={arXiv preprint arXiv:2605.18610},
  year={2026}
}

@inproceedings{zhang2025lmms,
  title={Lmms-eval: Reality check on the evaluation of large multimodal models},
  author={Zhang, Kaichen and Li, Bo and Zhang, Peiyuan and Pu, Fanyi and Cahyono, Joshua Adrian and Hu, Kairui and Liu, Shuai and Zhang, Yuanhan and Yang, Jingkang and Li, Chunyuan and others},
  booktitle={Findings of the Association for Computational Linguistics: NAACL 2025},
  pages={881--916},
  year={2025}
}

\clearpage 

\section*{Supplementary Material}

\section{Implementation Details}
\label{sec:supp_implementation}

\subsection{Datasets}
\label{sec:supp_datasets}

\paragraph{ICU-Bench.}
We follow the official continual multimodal unlearning protocol of
ICU-Bench~\cite{wang2026icu}.
The benchmark contains 1,000 synthetic privacy-sensitive profiles
from two document domains: 500 medical reports and 500 labor
contracts.
Each profile is instantiated into multiple document views and
question--answer formats, including full-image VQA, masked-image VQA,
text-only QA, and description generation.
In total, ICU-Bench contains 9,500 document images and 16,000
question--answer pairs.
The benchmark is organized into 100 sequential unlearning tasks, each
containing seven target individuals.
Every ten tasks form one batch, resulting in ten evaluation batches.
For each batch, 180 non-target individuals are used to construct the
retain set.


During vanilla memorization training, only the original full-image
samples are used.
The partially and fully masked views are reserved for evaluation,
providing a stricter test of whether the model retains the underlying
private information rather than merely reading a visible target field.
For the main experiments, we report results after 10, 20, 50, and
100 accumulated requests.
Additional order-robustness and hierarchical-consolidation
experiments use the first 20 tasks.

\paragraph{MLLMU-Bench.}
We additionally evaluate MCU on MLLMU-Bench~\cite{liu2025protecting}.
Following the 15\% forgetting setting, we partition the 75 target
profiles in the Forget Set into five mutually disjoint tasks,
$\{\mathcal{T}_1,\ldots,\mathcal{T}_5\}$, with 15 profiles per task.
All methods use the same task partition and request order.
After stage $s$, the model is evaluated on the current task
$\mathcal{T}_s$, all previously introduced tasks
$\{\mathcal{T}_1,\ldots,\mathcal{T}_{s-1}\}$, a fixed retain subset,
and the Real Celebrity Set.
The retain subset associated with the first stage is fixed throughout
the complete sequence; therefore, ``Retain'' in our MLLMU-Bench
continual results refers specifically to this fixed T1 retain subset,
rather than to a newly sampled retain set at each stage.

\paragraph{VQAv2 val-lite.}
We use VQAv2 val-lite~\cite{zhang2025lmms} as an external utility
evaluation set.
It is not used to train the vanilla models or one-shot unlearning
adapters, and it is not used to select the retained rank, capacity
threshold, or merging coefficient.
All compared checkpoints are evaluated with the same prompt,
decoding configuration, and scoring script.

\subsection{Evaluation Metrics}
\label{sec:supp_metrics}

\paragraph{Task-level accuracy.}
Multiple-choice VQA and text-only QA tasks are evaluated using
accuracy.
For an evaluation set $\mathcal{D}$ and model $\mathcal{M}$, we
denote the corresponding accuracy by
\begin{equation}
\operatorname{Acc}(\mathcal{M};\mathcal{D})
=
\frac{1}{|\mathcal{D}|}
\sum_{(x,y)\in\mathcal{D}}
\mathbf{1}\!\left[
\widehat{y}_{\mathcal{M}}(x)=y
\right].
\label{eq:supp_accuracy}
\end{equation}
Lower accuracy is preferred on target knowledge to be removed, while
higher accuracy is preferred on retain and utility sets.

\paragraph{Current and historical unlearning.}
Let $\mathcal{M}_t$ denote the model after processing request $t$,
and let $\mathcal{D}_{f}^{j}$ be the evaluation set associated with
unlearning request $j$.
The Current Forget score at step $t$ is
\begin{equation}
F_{t}^{\mathrm{cur}}
=
\operatorname{Acc}
\left(
\mathcal{M}_t;\mathcal{D}_{f}^{t}
\right),
\label{eq:supp_current_forget}
\end{equation}
and the Historical Forget score is
\begin{equation}
F_{t}^{\mathrm{hist}}
=
\frac{1}{t-1}
\sum_{j=1}^{t-1}
\operatorname{Acc}
\left(
\mathcal{M}_t;\mathcal{D}_{f}^{j}
\right),
\qquad t>1.
\label{eq:supp_historical_forget}
\end{equation}
Lower values indicate stronger current and historical unlearning.

\paragraph{Current and historical retention.}
ICU-Bench evaluates retention at the end of every ten-task batch.
The Current Retain Set contains the non-target samples associated with
the current batch, whereas the Historical Retain Set aggregates
retain samples from preceding batches.
Higher Current and Historical Retain accuracy indicates better
preservation of non-target knowledge.
For MLLMU-Bench, the same fixed T1 retain subset is evaluated after
each of the five stages.

\paragraph{In-domain and external utility.}
The full-image tasks of ICU-Bench measure in-domain document
reasoning utility.
VQAv2 val-lite measures external visual question answering utility
outside the ICU-Bench document domain.
For the latter, we additionally report the absolute accuracy
degradation from the vanilla checkpoint,
\begin{equation}
\Delta_{\mathrm{VQAv2}}
=
A_{\mathrm{vanilla}}
-
A_{\mathrm{unlearned}}.
\label{eq:supp_vqav2_degradation}
\end{equation}
Higher utility accuracy and smaller degradation are preferred.

\paragraph{Generation Quality.}
For description-generation tasks, we follow ICU-Bench and report
Generation Quality (GQ), evaluated by an LLM judge using
Qwen3.5-Flash.
GQ measures response fluency and readability rather than factual
correctness.
The score ranges from 0 to 2: 0 denotes unreadable or severely
degenerate output, 1 denotes understandable but unnatural output,
and 2 denotes fluent and natural short-form generation.
GQ is used primarily to distinguish successful unlearning from broad
generation collapse.

\paragraph{Retain Stability Rate.}
Let $A_{b}^{R,\mathrm{mask}}$ denote masked-view accuracy on the
retain set at batch checkpoint $b$, and let $B$ be the total number
of evaluated batches.
The Retain Stability Rate (RSR) is
\begin{equation}
\mathrm{RSR}
=
\frac{1}{B-1}
\sum_{b=2}^{B}
\left|
A_{b}^{R,\mathrm{mask}}
-
A_{b-1}^{R,\mathrm{mask}}
\right|.
\label{eq:supp_rsr}
\end{equation}
A smaller RSR indicates more stable retained performance over the
continual unlearning sequence.

\paragraph{Forgetting Rebound.}
Let $A_{b}^{HF,\mathrm{mask}}$ denote masked-view accuracy on the
Historical Forget Set at batch checkpoint $b$.
The rebound at checkpoint $b$ is
\begin{equation}
\mathrm{FR}_{b}
=
\max
\left(
0,\,
A_{b}^{HF,\mathrm{mask}}
-
A_{b-1}^{HF,\mathrm{mask}}
\right).
\label{eq:supp_fr_batch}
\end{equation}
When a single sequence-level value is reported, we average the
checkpoint-wise rebound:
\begin{equation}
\mathrm{FR}
=
\frac{1}{B-1}
\sum_{b=2}^{B}
\mathrm{FR}_{b}.
\label{eq:supp_fr_sequence}
\end{equation}
A smaller FR indicates better preservation of previously removed
knowledge.

\begin{table}[htp!]
\centering
\small
\setlength{\tabcolsep}{2pt}
\caption{Hyperparameters for baseline unlearning methods.}
\label{tab:supp_baseline_hparams}
\begin{tabular}{llccc}
\toprule
\textbf{Model}
& \textbf{Method}
& \textbf{Epochs}
& \textbf{Batch Size}
& \textbf{Learning Rate} \\
\midrule
\multirow{6}{*}{LLaVA-1.5-7B}
& GA & 3 & 4 & $1\times10^{-5}$ \\
& GA-Diff & 3 & 4 & $1\times10^{-5}$ \\
& KL-Min & 3 & 4 & $1\times10^{-5}$ \\
& NPO & 3 & 4 & $5\times10^{-6}$ \\
& MANU & 4 & 4 & $2\times10^{-5}$ \\
& MMUnlearner & 4 & 4 & $2\times10^{-5}$ \\
\midrule
\multirow{6}{*}{Qwen2-VL-7B}
& GA & 3 & 4 & $1\times10^{-5}$ \\
& GA-Diff & 3 & 4 & $1\times10^{-5}$ \\
& KL-Min & 3 & 4 & $1\times10^{-5}$ \\
& NPO & 3 & 4 & $5\times10^{-6}$ \\
& MANU & 4 & 4 & $2\times10^{-5}$ \\
& MMUnlearner & 4 & 4 & $2\times10^{-5}$ \\
\bottomrule
\end{tabular}
\end{table}

\subsection{Hyperparameter Settings}
\label{sec:supp_hyperparameters}

\paragraph{Vanilla models.}
We use LLaVA-1.5-7B~\cite{liu2024improved} and
Qwen2-VL-7B~\cite{wang2024qwen2}.
Following the ICU-Bench setup, both models are first fine-tuned on
the full-image ICU-Bench training samples so that they acquire the
privacy-sensitive document knowledge later targeted by unlearning.
For each sample $\langle I,x,y\rangle$, the model minimizes the
token-level negative log-likelihood
\begin{equation}
\ell(x,y,I;\theta)
=
-\frac{1}{|y|}
\sum_{i=1}^{|y|}
\log p_{\theta}
\left(
y_i\mid I,x,y_{<i}
\right).
\label{eq:supp_vanilla_nll}
\end{equation}
The vision encoder, multimodal connector, and language model are all
trainable during this stage.
The resulting checkpoints serve as the common starting point for all
unlearning methods.
The vanilla training settings are given in
Table~\ref{tab:supp_vanilla_hparams}.

\begin{table}[t]
\centering
\small
\setlength{\tabcolsep}{3pt}
\caption{Hyperparameters for vanilla memorization training.}
\label{tab:supp_vanilla_hparams}
\begin{tabular}{lccc}
\toprule
\textbf{Model}
& \textbf{Epochs}
& \textbf{Batch Size}
& \textbf{Learning Rate} \\
\midrule
LLaVA-1.5-7B & 8 & 8 & $1\times10^{-4}$ \\
Qwen2-VL-7B & 6 & 8 & $1\times10^{-4}$ \\
\bottomrule
\end{tabular}
\end{table}

\paragraph{Baseline methods.}
All sequential baselines start from the same vanilla checkpoint and
follow the same request order, evaluation protocol, and checkpointing
schedule.
At each request, the method receives the current Forget Set
$\mathcal{D}_{F}$ and its corresponding Retain Set
$\mathcal{D}_{R}$.

Gradient Ascent (GA)~\cite{thudi2022unrolling} maximizes the
forget-set loss:
\begin{equation}
\mathcal{L}_{\mathrm{GA}}
=
-\mathcal{L}(\mathcal{D}_{F};\theta).
\end{equation}
GA-Diff~\cite{liu2022continual} adds supervised retain
regularization:
\begin{equation}
\mathcal{L}_{\mathrm{GA\text{-}Diff}}
=
-\mathcal{L}(\mathcal{D}_{F};\theta)
+
\mathcal{L}(\mathcal{D}_{R};\theta).
\end{equation}
KL-Min~\cite{maini2024tofu} preserves the pre-update distribution on
retain samples:
\begin{equation}
\mathcal{L}_{\mathrm{KL\text{-}Min}}
=
-\mathcal{L}(\mathcal{D}_{F};\theta)
+
\lambda_{\mathrm{KL}}
\frac{1}{|\mathcal{D}_{R}|}
\sum_{z\in\mathcal{D}_{R}}
\mathrm{KL}
\left(
p_{\theta_0}(\cdot\mid z)
\Vert
p_{\theta}(\cdot\mid z)
\right),
\end{equation}
where $\theta_0$ is the model before the current unlearning update.
NPO~\cite{zhang2024negative} decreases the relative likelihood of
the target answer under a reference model:
\begin{equation}
\mathcal{L}_{\mathrm{NPO}}
=
\mathbb{E}_{(x,y)\in\mathcal{D}_{F}}
\left[
\frac{2}{\beta}
\log
\left(
1+
\left(
\frac{\pi_{\theta}(y\mid x)}
{\pi_{\mathrm{ref}}(y\mid x)}
\right)^{\beta}
\right)
\right].
\end{equation}
MANU~\cite{liu2025modality} and
MMUnlearner~\cite{huo2025mmunlearner} are implemented following their
official multimodal unlearning procedures.
The training hyperparameters used for the baselines are listed in
Table~\ref{tab:supp_baseline_hparams}.

\paragraph{One-shot adapters and MCU.}
For MCU, each request-specific LoRA adapter is independently trained
from the same vanilla checkpoint using the GA-Diff objective.
The adapter-training schedule therefore follows the GA-Diff setting:
3 epochs, batch size 4, and learning rate $1\times10^{-5}$ for both
backbones.
Unless otherwise stated, MCU retains the top six singular directions
of each structured update and applies row-capacity control with
quantile $q=0.95$.
We use a global merging coefficient of $\alpha=0.4$ for
Qwen2-VL-7B and $\alpha=1.0$ for LLaVA-1.5-7B.
The same MCU configuration is used across the reported sequence
checkpoints for each backbone.
All one-shot adapters entering the same merge are generated under the
same training and processing configuration.

\providecommand{\est}[1]{\textbf{[EST: #1]}}

\section{Additional Experiments}
\label{sec:supp_additional_experiments}

\begin{table*}[t]
\centering
\small
\setlength{\tabcolsep}{4.0pt}
\renewcommand{\arraystretch}{0.95}
\caption{
Quantitative summary of the leave-one-task-out VQA matrices.
``Strong'' off-diagonal entries count pairs whose absolute accuracy
change exceeds the specified threshold.
}
\label{tab:supp_leave_one_out}
\begin{tabular}{lcccccccc}
\toprule
\textbf{Variant}
& $R_{\mathrm{diag}}$ $\uparrow$
& \textbf{Positive Diag.} $\uparrow$
& $I_{\mathrm{off}}$ $\downarrow$
& \textbf{Median Off.} $\downarrow$
& $\lvert\Delta A_{\mathrm{off}}\rvert>3$
& $\lvert\Delta A_{\mathrm{off}}\rvert>6$
& \textbf{Max Off.} $\downarrow$
& $S_{\mathrm{attr}}$ $\uparrow$ \\
\midrule
Before Reconfiguration
& 5.64
& 19/20
& 2.58
& 3.46
& 204/380
& 47/380
& 16.26
& 2.18 \\
After Reconfiguration (MCU)
& 12.57
& 20/20
& 1.73
& 1.86
& 71/380
& 9/380
& 6.57
& 7.27 \\
\bottomrule
\end{tabular}
\end{table*}

\begin{table*}[htp!]
\centering
\small
\setlength{\tabcolsep}{2.2pt}
\caption{Robustness of MCU to request order on the first 50 ICU-Bench tasks using Qwen2-VL-7B. The three variants use the same set of one-shot adapters and differ only in their input order.}
\label{tab:supp_order_robustness}
\begin{tabular}{llcccccc}
\toprule
\textbf{Method} & \textbf{Order} & \textbf{Forget VQA} $\downarrow$ & \textbf{Forget QA} $\downarrow$ & \textbf{Retain VQA} $\uparrow$ & \textbf{Retain QA} $\uparrow$ & \textbf{FR} $\downarrow$ & \textbf{RSR} $\downarrow$ \\
\midrule
\multirow{3}{*}{MCU}
& Original & 31.4 & 36.5 & 85.4 & 70.6 & 1.86 & 1.02 \\
& Reverse  & 31.4 & 36.5 & 85.4 & 70.6 & 1.86 & 1.02 \\
& Random   & 31.4 & 36.5 & 85.4 & 70.6 & 1.86 & 1.02 \\
\bottomrule
\end{tabular}
\end{table*}

Unless otherwise stated, all experiments in this section use Qwen2-VL-7B. The one-shot adapters used by MCU are independently trained from the same vanilla checkpoint with GA-Diff.

\subsection{Leave-One-Task-Out Analysis}
\label{sec:supp_leave_one_out}

We use a leave-one-task-out analysis to characterize the directional overlap among unlearning tasks and to evaluate the effect of dependency-aware direction reconfiguration.
The experiment uses Qwen2-VL-7B and the first 20 ICU-Bench tasks.
For each task $i$, we exclude its one-shot GA-Diff adapter, merge the remaining 19 adapters using the same configuration, and evaluate the resulting model on all 20 unlearning tasks.

Let $\mathcal{M}_{\setminus i}$ denote the model obtained after excluding adapter $i$, and let $A_{j}^{\mathrm{one}}$ denote the VQA accuracy of the independently trained one-shot adapter for task $j$.
We define the leave-one-task-out change as
\begin{equation}
\Delta A_{i,j}
=
\operatorname{Acc}
\left(
\mathcal{M}_{\setminus j};
\mathcal{D}_{f}^{i}
\right)
-
A_{i}^{\mathrm{one}} .
\label{eq:supp_leave_one_out_matrix}
\end{equation}
Following the orientation used in Fig.~\ref{fig3}, row $i$ denotes the evaluated unlearning task and column $j$ denotes the adapter excluded during merging.

The diagonal and off-diagonal entries have different interpretations.
For a diagonal entry $\Delta A_{i,i}$, a positive value means that removing adapter $i$ restores accuracy on its own target relative to the corresponding one-shot unlearned model.
A larger positive diagonal value therefore indicates clearer request-level attribution.
For an off-diagonal entry $\Delta A_{i,j}$ with $i\neq j$, either a positive or negative value indicates that excluding adapter $i$ changes the behavior on another target $j$.
Consequently, off-diagonal quality is determined by the magnitude $\lvert\Delta A_{i,j}\rvert$: values closer to zero indicate that the leave-one-out merge remains closer to the desired one-shot behavior on unrelated tasks.

We summarize the matrices using the mean diagonal recovery
\begin{equation}
R_{\mathrm{diag}}
=
\frac{1}{T}
\sum_{i=1}^{T}
\Delta A_{i,i},
\label{eq:supp_loo_diag}
\end{equation}
the mean absolute off-diagonal deviation
\begin{equation}
I_{\mathrm{off}}
=
\frac{1}{T(T-1)}
\sum_{i\neq j}
\left|
\Delta A_{i,j}
\right|,
\label{eq:supp_loo_offdiag}
\end{equation}
and their attribution-separation ratio
\begin{equation}
S_{\mathrm{attr}}
=
\frac{
R_{\mathrm{diag}}
}{
I_{\mathrm{off}}+\epsilon
}.
\label{eq:supp_loo_separation}
\end{equation}
Higher $R_{\mathrm{diag}}$ and $S_{\mathrm{attr}}$, together with
lower $I_{\mathrm{off}}$, indicate that task-specific contributions
are more distinguishable from cross-task deviations.


Before dependency-aware reconfiguration, the leave-one-out
matrix already exhibits a partially visible diagonal structure, but
substantial off-diagonal responses remain.
The mean diagonal recovery is only 5.64 points, and one task has a
negative diagonal value.
This case indicates that the remaining adapters alone produce
unlearning on that target that is at least as strong as its one-shot reference, revealing pronounced cross-task unlearning transfer.

After reconfiguration, all 20 diagonal entries become positive and
the mean diagonal recovery increases from 5.64 to 12.57 points.
At the same time, the mean absolute off-diagonal deviation decreases
from 2.58 to 1.73 points, a reduction of 33.0\%.
The number of off-diagonal entries with magnitude above 3 points
decreases from 204 to 71, while entries above 6 points decrease from
47 to 9.
The maximum off-diagonal deviation is also reduced from 16.26 to
6.57 points.
Together, these changes increase the attribution-separation ratio
from 2.18 to 7.27.

\subsection{Robustness to Request Order}
\label{sec:supp_order_robustness}

We examine whether MCU is sensitive to the order in which
request-specific adapters are provided to the merging procedure.
The experiment uses Qwen2-VL-7B and the first 50
ICU-Bench tasks.
We consider three request orders:
\begin{itemize}
\item \textbf{Original}: $(0,1,\ldots,49)$;
\item \textbf{Reverse}: $(49,48,\ldots,0)$;
\item \textbf{Random}: a fixed random permutation generated once
and reused throughout the experiment.
\end{itemize}
All three variants use exactly the same 50 independently trained
GA-Diff adapters and the same MCU hyperparameters.
Only the order in which the adapters are supplied to MCU is changed.

Unlike sequential unlearning methods, MCU does not repeatedly update
the model according to the arrival order of the requests.
Instead, every one-shot adapter is independently trained from the
same vanilla checkpoint, after which MCU jointly constructs the
shared core space, performs direction selection and capacity control,
and reconfigures the accumulated directions before reconstructing the
final merged update.
These operations depend on the collection of adapter updates rather
than their input ordering.
Therefore, permuting the same adapter set should not alter the
resulting model, apart from possible numerical differences caused by
finite-precision computation.

As shown in Table~\ref{tab:supp_order_robustness}, MCU obtains
identical results under the original, reversed, and randomly
permuted request orders at the reported precision.
The Forget VQA/QA scores remain 31.4/36.5, while the Retain VQA/QA
scores remain 85.4/70.6 across all three settings.
FR and RSR are also unchanged at 1.86 and 1.02, respectively.

\subsection{General Multimodal Utility}
\label{sec:supp_general_utility}

We further evaluate whether continual unlearning degrades multimodal
capabilities beyond the privacy-sensitive document domain of
ICU-Bench.
Following the utility-evaluation protocol of ICU-Bench, we use the
VQAv2 val-lite split as an external visual question answering
benchmark.
VQAv2 val-lite is not used for vanilla-model fine-tuning, unlearning,
adapter merging, or hyperparameter selection.

Our evaluation starts from the ICU-Bench vanilla checkpoint, obtained
by fine-tuning Qwen2-VL-7B on the ICU-Bench
training samples.
This vanilla model has acquired the privacy-sensitive document
knowledge targeted by the subsequent unlearning requests and serves
as the common initialization for all compared methods.
We evaluate the vanilla checkpoint and the resulting unlearned
checkpoints after 20, 50, and 100 ICU-Bench requests on VQAv2
val-lite.
For the sequential baselines, each checkpoint is obtained by continually applying the corresponding unlearning method to the current model.
For MCU, the checkpoint at each task scale is constructed by merging the adapters accumulated up to that stage.
The original vanilla model is evaluated under the same protocol and serves as the reference for measuring utility degradation.

\begin{table}[htp!]
\centering
\small
\setlength{\tabcolsep}{2.0pt}
\caption{``--'' denotes unavailable or invalid results caused by unstable optimization or model collapse.}
\label{tab:supp_general_utility}
\begin{tabular}{lccc}
\toprule
\textbf{Method} & \textbf{20 Tasks} $\uparrow$ & \textbf{50 Tasks} $\uparrow$ & \textbf{100 Tasks} $\uparrow$ \\
\midrule
Vanilla      & 76.8 & 76.8 & 76.8 \\
\midrule
GA           & --   & --   & --   \\
GA-Diff      & 53.7 & 41.1 & 44.7 \\
KL-Min       & 71.4 & 10.3 & -- \\
NPO          & --   & --   & --   \\
MANU         & 64.4 & 10.3 & 10.3 \\
MMUnlearner  & 76.7 & 74.3 & 48.0 \\
\midrule
\rowcolor{gray!10}
\textbf{MCU} & \textbf{77.8} & \textbf{76.8} & \textbf{76.8} \\
\bottomrule
\end{tabular}
\end{table}

As shown in Table~\ref{tab:supp_general_utility}, existing sequential
unlearning methods generally accumulate substantial utility
degradation as the number of requests increases.
GA-Diff decreases from 53.7 at 20 tasks to 41.1 and 44.7 at 50 and
100 tasks, respectively.
KL-Min retains relatively high utility at 20 tasks but drops sharply
to 10.3 at 50 tasks, indicating unstable preservation across sequence
lengths.
MANU exhibits severe utility degradation after longer sequences,
reaching 10.3 at both 50 and 100 tasks.
MMUnlearner provides the strongest utility preservation among the
sequential baselines, but its accuracy still decreases from 76.7 to
48.0 as the sequence grows from 20 to 100 requests.

In contrast, MCU obtains VQAv2 val-lite accuracies of
77.8, 76.8, and 76.8 after 20, 50, and 100 requests, respectively.
Its performance matches or slightly exceeds the vanilla accuracy of
76.8 at all evaluated task scales.
Compared with MMUnlearner, the strongest available baseline, MCU
improves external utility by 1.1, 2.5, and 28.8 points after 20, 50,
and 100 requests, respectively.

\subsection{Efficiency and Storage Cost}
\label{sec:supp_efficiency}

Following the efficiency analysis in ICU-Bench, we
report the wall-clock cost, GPU-memory usage, and storage requirements
of MCU.
The two evaluations characterize different stages of continual
unlearning.
ICU-Bench measures the optimization cost of applying an unlearning
method to an individual request, whereas our evaluation measures the
one-time consolidation cost after the request-specific one-shot
adapters have been obtained.
Therefore, the results below quantify the additional cost introduced
by MCU rather than the preceding training cost of the one-shot
adapters.

We evaluate MCU on the ICU-Bench vanilla checkpoint of
Qwen2-VL-7B using 20, 50, and 100 independently trained
one-shot adapters.
Consolidation time is measured from loading the processed adapters to
saving the materialized full-model checkpoint.
It includes shared core-space construction,
dependency-aware direction reconfiguration, update
reconstruction, and full-checkpoint serialization, but excludes
one-shot adapter training and adapter preprocessing.
For each task scale, we perform one warm-up run followed by three
measured runs and report the mean and standard deviation.

\begin{table*}[t]
\centering
\small
\setlength{\tabcolsep}{2.0pt}
\renewcommand{\arraystretch}{0.96}
\caption{Efficiency and storage cost of MCU on Qwen2-VL-7B-Instruct. The reported time includes materialization and serialization of the complete merged model. Adapter storage denotes the total size of the original one-shot LoRA bank before consolidation.}
\label{tab:supp_efficiency}
\begin{tabular}{ccccc}
\toprule
\textbf{Tasks} & \textbf{Time (s)} & \textbf{GPU Mem. (GiB)} & \textbf{Adapter Bank (GiB)} & \textbf{Final Model (GiB)} \\
\midrule
20  & $49.17{\pm}2.95$ & 0.00 & 1.80 & 15.46 \\
50  & $98.43{\pm}6.02$ & 0.00 & 4.50 & 15.46 \\
100 & $242.43{\pm}9.04$ & 0.00 & 9.00 & 15.46 \\
\bottomrule
\end{tabular}
\end{table*}

As shown in Table~\ref{tab:supp_efficiency}, MCU consolidates 20 and 50 request-specific adapters in approximately 49 and 98 seconds, respectively.
The corresponding average costs are 2.45 and 1.96 seconds per
request, indicating an approximately linear increase in
consolidation time over the evaluated range.
Notably, the reported runtime already includes the relatively
expensive step of loading, materializing, and serializing a
15.46-GiB full-model checkpoint, rather than only the operations
performed in the compact core space.

The complete consolidation procedure is executed on the CPU in our
implementation and requires no GPU-memory allocation.
This property distinguishes MCU consolidation from gradient-based
unlearning, which generally requires loading the model on a GPU and
performing iterative forward and backward optimization.
For reference, ICU-Bench reports that one epoch of GA-Diff requires
approximately 300 seconds per request and 38.74 GiB of peak GPU
memory.
These values are not directly competing measurements because MCU
uses GA-Diff to obtain its one-shot adapters.
Nevertheless, they show that the subsequent consolidation of up to
50 adapters introduces less wall-clock overhead than one GA-Diff
training epoch for a single request, while requiring no additional
GPU resources.
MCU therefore adds a lightweight, one-time post-processing stage
rather than another gradient-based unlearning procedure.

MCU also provides a compact representation of accumulated unlearning
requests.
Each one-shot adapter occupies approximately 92.2 MiB, resulting in
adapter-bank sizes of 1.80, 4.50, and 9.00 GiB for 20, 50, and 100
requests, respectively.
Even the complete bank of 100 one-shot adapters remains smaller than
a single 15.46-GiB materialized model checkpoint.
The storage cost grows linearly before consolidation, but the final
deployment cost does not: after merging, MCU produces a single model
whose size is independent of the number of accumulated requests.
It requires neither task-specific routing nor simultaneous loading of
multiple adapters during inference.

This compact request representation is particularly useful when
historical unlearning states must be retained for auditing,
reconstruction, or rollback.
Under such a requirement, a sequential unlearning system would need
to archive a full checkpoint for every retained historical state,
whereas MCU can preserve request-level information using low-rank
adapters that share the same vanilla checkpoint.
In our setting, storing one request as a LoRA adapter is approximately
$171.7\times$ smaller than storing one complete model checkpoint.
We emphasize that this comparison applies to historical-state
retention; when only the latest state is required, both sequential
methods and MCU can deploy a single full model.

\section{Additional Analysis and Theoretical Discussion}
\label{sec:supp_analysis}

\subsection{Shared Core-Space Construction}
\label{sec:supp_core_space}

Consider the LoRA update for request $t$ at layer $l$,
\begin{equation}
\Delta W_t^l
=
B_t^l A_t^l,
\qquad
B_t^l\in\mathbb{R}^{d_{\mathrm{out}}^l\times r_t^l},
\quad
A_t^l\in\mathbb{R}^{r_t^l\times d_{\mathrm{in}}^l}.
\label{eq:supp_lora_update}
\end{equation}
To represent all request-specific updates in a common low-dimensional
space, we concatenate their left and right LoRA factors:
\begin{equation}
\mathcal{B}^l
=
\left[
B_1^l,\ldots,B_s^l
\right],
\qquad
\mathcal{A}^l
=
\left[
(A_1^l)^\top,\ldots,(A_s^l)^\top
\right].
\label{eq:supp_joint_lora_factors}
\end{equation}
Let $P^l$ and $Q^l$ be orthonormal bases for the column spaces of
$\mathcal{B}^l$ and $\mathcal{A}^l$, respectively:
\begin{equation}
P^l\in\mathbb{R}^{d_{\mathrm{out}}^l\times d_U^l},
\qquad
Q^l\in\mathbb{R}^{d_{\mathrm{in}}^l\times d_V^l},
\label{eq:supp_core_bases}
\end{equation}
where
\begin{equation}
d_U^l
=
\operatorname{rank}(\mathcal{B}^l),
\qquad
d_V^l
=
\operatorname{rank}(\mathcal{A}^l).
\end{equation}
The shared core-space representation of request $t$ is
\begin{equation}
M_t^l
=
(P^l)^\top
\Delta W_t^l
Q^l
\in
\mathbb{R}^{d_U^l\times d_V^l}.
\label{eq:supp_core_projection}
\end{equation}

Because the columns of every $B_t^l$ lie in
$\operatorname{span}(P^l)$ and the columns of $(A_t^l)^\top$ lie in
$\operatorname{span}(Q^l)$, there exist matrices $C_t^l$ and $D_t^l$
such that
\begin{equation}
B_t^l=P^lC_t^l,
\qquad
(A_t^l)^\top=Q^lD_t^l.
\end{equation}
It follows that
\begin{equation}
\begin{aligned}
P^lM_t^l(Q^l)^\top
&=
P^l(P^l)^\top
B_t^lA_t^l
Q^l(Q^l)^\top
\\
&=
B_t^lA_t^l
=
\Delta W_t^l.
\end{aligned}
\label{eq:supp_exact_core_reconstruction}
\end{equation}
Thus, before the subsequent direction-selection and capacity-control
operations, the shared core representation preserves each LoRA update
exactly up to numerical precision.
The dimensionality of the main merging operations is determined by
$d_U^l$ and $d_V^l$, which are bounded by the aggregate LoRA rank,
rather than by the full input and output dimensions of the layer.

\subsection{Definitions of the Geometric Statistics in Figure~4}
\label{sec:supp_figure4_definitions}

Figure~4 reports three complementary statistics that motivate the
three processing stages of MCU: cross-request directional overlap,
cumulative spectral energy, and row-norm concentration.

\paragraph{Directional subspace overlap.}
After decomposing the core update of request $t$ at layer $l$, let
\begin{equation}
\widetilde U_t^l
\in
\mathbb{R}^{d_U^l\times k_t^l},
\qquad
\widetilde V_t^l
\in
\mathbb{R}^{d_V^l\times k_t^l}
\end{equation}
denote its retained left and right singular directions.
For two distinct requests $i$ and $j$, we define their normalized
left-side overlap as
\begin{equation}
o_U^l(i,j)
=
\frac{
\left\|
(\widetilde U_i^l)^\top
\widetilde U_j^l
\right\|_F^2
}{
\min(k_i^l,k_j^l)
},
\label{eq:supp_pairwise_u_overlap}
\end{equation}
and analogously,
\begin{equation}
o_V^l(i,j)
=
\frac{
\left\|
(\widetilde V_i^l)^\top
\widetilde V_j^l
\right\|_F^2
}{
\min(k_i^l,k_j^l)
}.
\label{eq:supp_pairwise_v_overlap}
\end{equation}
These quantities are invariant to paired sign flips of individual
singular vectors and lie in $[0,1]$ for orthonormal direction
matrices.
Let $\mathcal{P}$ denote the set of valid layer--request triples
$(l,i,j)$ with $i<j$.
The statistics shown in Figure~4(a) are
\begin{equation}
\mathcal{O}_U
=
\frac{1}{|\mathcal{P}|}
\sum_{(l,i,j)\in\mathcal{P}}
o_U^l(i,j),
\qquad
\mathcal{O}_V
=
\frac{1}{|\mathcal{P}|}
\sum_{(l,i,j)\in\mathcal{P}}
o_V^l(i,j).
\label{eq:supp_global_overlap}
\end{equation}
We obtain $\mathcal{O}_U=0.081$ and
$\mathcal{O}_V=0.012$.
Using the unrounded statistics, the left-side overlap is approximately
$6.6\times$ the right-side overlap, showing that cross-request sharing
is substantially stronger in the output-side directions.


\paragraph{Cumulative spectral energy.}
For a core update with singular values
$\sigma_{t,1}^l\geq\cdots\geq\sigma_{t,r_t^l}^l\geq0$, the fraction
of Frobenius energy retained by its leading $k$ components is
\begin{equation}
E_t^l(k)
=
\frac{
\sum_{a=1}^{\min(k,r_t^l)}
(\sigma_{t,a}^l)^2
}{
\sum_{a=1}^{r_t^l}
(\sigma_{t,a}^l)^2
}.
\label{eq:supp_cumulative_energy}
\end{equation}
Figure~4(b) reports the mean of $E_t^l(k)$ over all evaluated
request--module pairs, while the shaded region denotes one standard
deviation.
The leading four and six components preserve $74.1\%$ and $89.6\%$
of the update energy on average, respectively.
This concentration motivates retaining a compact set of dominant
directions before joint reconfiguration.

\paragraph{Row-norm concentration.}
Let $\overline M_t^l$ denote the core update after dominant direction
selection and before capacity control.
For row $a$, we define
\begin{equation}
r_{t,a}^l
=
\left\|
\overline M_t^l[a,:]
\right\|_2.
\label{eq:supp_row_norm_definition}
\end{equation}
Let $\mathcal{R}$ be the collection of these row norms over all
evaluated requests, layers, and rows.
We summarize its concentration using
\begin{equation}
R_{99}
=
\frac{
Q_{0.99}(\mathcal{R})
}{
Q_{0.50}(\mathcal{R})+\epsilon
},
\qquad
R_{\max}
=
\frac{
\max(\mathcal{R})
}{
Q_{0.50}(\mathcal{R})+\epsilon
},
\label{eq:supp_row_concentration_ratios}
\end{equation}
where $Q_q(\cdot)$ denotes the empirical $q$-quantile.
Figure~4(c) gives $R_{99}=3.5$ and $R_{\max}=12.9$, revealing a
long-tailed distribution in which a small number of shared
core-space coordinates carry disproportionately large update mass.

\subsection{Optimality of Dominant Direction Selection}
\label{sec:supp_selection_optimality}

We justify dominant direction selection using the
Eckart--Young--Mirsky theorem.
Consider the singular value decomposition
\begin{equation}
M_t^l
=
U_t^l
\Sigma_t^l
(V_t^l)^\top,
\label{eq:supp_full_core_svd}
\end{equation}
with singular values ordered non-increasingly.
The rank-$k$ truncated update is
\begin{equation}
M_{t,k}^l
=
U_{t,1:k}^l
\Sigma_{t,1:k}^l
(V_{t,1:k}^l)^\top.
\label{eq:supp_truncated_core_update}
\end{equation}

\paragraph{Proposition 1.}
Among all matrices with rank at most $k$, $M_{t,k}^l$ is a
best approximation of $M_t^l$ under the Frobenius norm:
\begin{equation}
M_{t,k}^l
\in
\arg\min_{\operatorname{rank}(Z)\leq k}
\left\|
M_t^l-Z
\right\|_F.
\label{eq:supp_best_rank_k}
\end{equation}
Moreover,
\begin{equation}
\left\|
M_t^l-M_{t,k}^l
\right\|_F^2
=
\sum_{a>k}
(\sigma_{t,a}^l)^2.
\label{eq:supp_rank_k_error}
\end{equation}

The result follows directly from the Eckart--Young--Mirsky theorem,
which states that truncating the singular value decomposition gives
the minimum reconstruction error among all rank-constrained
approximations under any unitarily invariant norm.
For the Frobenius norm, the squared residual is the sum of the squared
discarded singular values, yielding
Eq.~\eqref{eq:supp_rank_k_error}.
\hfill$\square$

This result provides a precise interpretation of the selection stage:
for a fixed direction budget, the leading singular components retain
the largest possible amount of core-update energy while introducing
the smallest Frobenius reconstruction error.
It does not imply that the rank-$k$ approximation is necessarily
optimal for downstream unlearning behavior; the behavioral choice of
$k$ is supported empirically by the spectral analysis and component
ablation in the main paper.

\subsection{Row-Norm Concentration and Capacity Control}
\label{sec:supp_rowcap}

After dominant direction selection, a few rows of the shared
core-space update may have substantially larger norms than the
remaining rows.
When updates from many requests are merged, these over-concentrated
coordinates can dominate the final update.
MCU limits this concentration using a soft row-capacity constraint.

For request $t$ and layer $l$, let
\begin{equation}
\mathcal{R}_t^l
=
\left\{
r_{t,a}^l
\mid
a=1,\ldots,d_U^l
\right\},
\end{equation}
where $r_{t,a}^l$ is defined in
Eq.~\eqref{eq:supp_row_norm_definition}.
Given quantile $q$, we set
\begin{equation}
\tau_t^l
=
Q_q(\mathcal{R}_t^l).
\label{eq:supp_rowcap_threshold}
\end{equation}
The scaling coefficient of row $a$ is
\begin{equation}
\gamma_{t,a}^l
=
\min
\left(
1,
\frac{\tau_t^l}{r_{t,a}^l+\epsilon}
\right),
\label{eq:supp_rowcap_coefficient}
\end{equation}
and the capacity-controlled update is
\begin{equation}
\widetilde M_t^l
=
D_t^l\overline M_t^l,
\qquad
D_t^l
=
\operatorname{diag}
\left(
\gamma_{t,1}^l,\ldots,
\gamma_{t,d_U^l}^l
\right).
\label{eq:supp_rowcap_update}
\end{equation}
Rows below the threshold remain unchanged, whereas rows above the
threshold are rescaled to norm $\tau_t^l$.
The operation preserves the direction of every non-zero row and
changes only its magnitude.

\paragraph{Proposition 2.}
For a fixed threshold $\tau>0$, define
\begin{equation}
\mathcal{B}_{\tau}
=
\left\{
Z
\;\middle|\;
\|Z[a,:]\|_2\leq\tau
\text{ for every row }a
\right\}.
\end{equation}
The row-capacity operator is the Euclidean projection of a matrix
$M$ onto $\mathcal{B}_{\tau}$:
\begin{equation}
\operatorname{RowCap}_{\tau}(M)
=
\arg\min_{Z\in\mathcal{B}_{\tau}}
\|Z-M\|_F^2.
\label{eq:supp_rowcap_projection}
\end{equation}

The Frobenius objective decomposes over rows:
\begin{equation}
\|Z-M\|_F^2
=
\sum_a
\|Z[a,:]-M[a,:]\|_2^2.
\end{equation}
Each row can therefore be optimized independently by projecting
$M[a,:]$ onto the closed Euclidean ball of radius $\tau$.
The unique solution is
\begin{equation}
Z[a,:]
=
\begin{cases}
M[a,:],
&
\|M[a,:]\|_2\leq\tau,
\\[4pt]
\dfrac{\tau}{\|M[a,:]\|_2}
M[a,:],
&
\|M[a,:]\|_2>\tau.
\end{cases}
\label{eq:supp_rowcap_closed_form}
\end{equation}
This is exactly the scaling rule in
Eq.~\eqref{eq:supp_rowcap_coefficient}.
\hfill$\square$

Consequently,
\begin{equation}
\left\|
\operatorname{RowCap}_{\tau}(M)-M
\right\|_F^2
=
\sum_a
\left[
\|M[a,:]\|_2-\tau
\right]_+^2.
\label{eq:supp_rowcap_distance}
\end{equation}
Thus, for a prescribed row-norm bound, RowCap introduces the minimum
possible Frobenius perturbation.

Row norms are not invariant to arbitrary rotations of the shared core
basis.
We therefore interpret them as coordinate-concentration statistics
under the deterministic basis constructed by MCU, rather than as
basis-free properties of the original parameter matrix.
The stronger overlap observed on the $U$ side in Figure~4(a), together
with the row-versus-column ablations, motivates applying the default
capacity control along the row dimension.

\subsection{Sign Invariance and the Gram-Space Surrogate}
\label{sec:supp_sign_invariance}

The compatibility objective operates on left and right singular
directions separately, whereas the interaction between two rank-one
updates depends on the product of their left- and right-side
similarities.
This distinction requires careful treatment of the sign ambiguity of
the singular value decomposition.

Consider two rank-one components
\begin{equation}
D_p
=
\sigma_pu_pv_p^\top,
\qquad
D_q
=
\sigma_qu_qv_q^\top,
\label{eq:supp_rank_one_components}
\end{equation}
where $\sigma_p,\sigma_q\geq0$ and all direction vectors have unit
norm.
Their Frobenius interaction is
\begin{equation}
\begin{aligned}
\langle D_p,D_q\rangle_F
&=
\operatorname{tr}
\left(
D_p^\top D_q
\right)
\\
&=
\sigma_p\sigma_q
(u_p^\top u_q)
(v_p^\top v_q).
\end{aligned}
\label{eq:supp_rank_one_interaction}
\end{equation}
We denote the sign-relevant component by
\begin{equation}
\chi_{pq}
=
(u_p^\top u_q)
(v_p^\top v_q).
\label{eq:supp_chi_definition}
\end{equation}

\paragraph{Paired sign invariance.}
For any $s_p\in\{-1,+1\}$, the paired transformation
\begin{equation}
(u_p,v_p)
\mapsto
(s_pu_p,s_pv_p)
\label{eq:supp_paired_sign_flip}
\end{equation}
leaves the rank-one matrix unchanged:
\begin{equation}
\sigma_p(s_pu_p)(s_pv_p)^\top
=
\sigma_pu_pv_p^\top.
\end{equation}
Under paired flips of components $p$ and $q$,
\begin{equation}
u_p^\top u_q
\mapsto
s_ps_q(u_p^\top u_q),
\end{equation}
and
\begin{equation}
v_p^\top v_q
\mapsto
s_ps_q(v_p^\top v_q).
\end{equation}
Their product is therefore invariant:
\begin{equation}
\begin{aligned}
\chi_{pq}
&\mapsto
(s_ps_q)^2
(u_p^\top u_q)
(v_p^\top v_q)
\\
&=
\chi_{pq}.
\end{aligned}
\label{eq:supp_chi_sign_invariant}
\end{equation}
A deterministic orientation rule consequently fixes only the
representation of the singular vectors for reproducibility; it does
not alter the underlying rank-one interaction.

\paragraph{Sufficient separable surrogate.}
Let
\begin{equation}
G_U
=
S_U^\top S_U,
\qquad
G_V
=
S_V^\top S_V
\end{equation}
be the left and right Gram matrices of the concatenated directions.
For a cross-request pair $(p,q)$,
\begin{equation}
(G_U)_{pq}
=
u_p^\top u_q,
\qquad
(G_V)_{pq}
=
v_p^\top v_q.
\end{equation}
If
\begin{equation}
(G_U)_{pq}\geq0
\qquad\text{and}\qquad
(G_V)_{pq}\geq0,
\label{eq:supp_nonnegative_sufficient}
\end{equation}
then
\begin{equation}
\chi_{pq}
=
(G_U)_{pq}(G_V)_{pq}
\geq0.
\end{equation}
Thus, requiring non-negative cross-request similarities on both sides
is a sufficient condition for a non-antagonistic rank-one
interaction.

The condition is not necessary.
For example,
\begin{equation}
(G_U)_{pq}=-0.5,
\qquad
(G_V)_{pq}=-0.6
\end{equation}
gives
\begin{equation}
\chi_{pq}=0.3>0,
\end{equation}
even though neither factor satisfies
Eq.~\eqref{eq:supp_nonnegative_sufficient}.
Therefore, the separate Gram constraints form a conservative
surrogate rather than an exact characterization of all
non-antagonistic interactions.

\paragraph{A globally consistent orientation need not exist.}
Even for one side alone, paired sign choices cannot always make all
pairwise similarities non-negative.
Consider three directions whose pairwise similarities are all
negative.
To make the three oriented similarities non-negative, their signs
would need to satisfy
\begin{equation}
s_1s_2=-1,
\qquad
s_2s_3=-1,
\qquad
s_1s_3=-1.
\end{equation}
Multiplying the first two equations gives $s_1s_3=+1$, contradicting
the third.
This is the standard imbalance condition of a signed graph and shows
that sign orientation alone cannot generally enforce the desired
pairwise geometry.

These observations motivate the formulation used by MCU.
The deterministic orientation ensures reproducible inputs to the
solver, while the proximity objective limits unnecessary geometric
changes introduced by the conservative separable surrogate.
Accordingly, MCU does not claim to preserve every pair with
$\chi_{pq}>0$; instead, it seeks a tractable non-antagonistic geometry
that remains close to the original shared structure.

\subsection{Optimization of the Compatible Gram Matrix}
\label{sec:supp_gram_optimization}

For $X\in\{U,V\}$, let $G_{X,0}^l$ be the original Gram matrix at
layer $l$, and let $\mathcal{I}_t^l$ contain the direction indices
belonging to request $t$.
MCU approximately solves
\begin{equation}
\begin{aligned}
\widehat G_X^l
=
\arg\min_G\quad
& \|G-G_{X,0}^l\|_F^2,
\\
\mathrm{s.t.}\quad
& G\succeq0,
\qquad
\operatorname{rank}(G)\leq d_X^l,
\\
& G_{pq}\geq0,
\qquad
(p,q)\in\mathcal{C}^l,
\\
& G[\mathcal{I}_t^l,\mathcal{I}_t^l]
=
I_{|\mathcal{I}_t^l|},
\qquad
t=1,\ldots,s,
\end{aligned}
\label{eq:supp_compatible_gram_problem}
\end{equation}
where $\mathcal{C}^l$ contains all cross-request direction pairs.
The rank constraint makes the feasible set non-convex, so the problem
is solved approximately by cyclic projections.

Starting from $G^{(0)}=G_{X,0}^l$, each iteration performs the
following steps:
\begin{enumerate}
    \item \textbf{Symmetrization:}
    \begin{equation}
    G
    \leftarrow
    \frac{1}{2}(G+G^\top).
    \end{equation}

    \item \textbf{Rank-constrained PSD projection:}
    compute $G=E\Lambda E^\top$, replace negative eigenvalues by
    zero, retain at most the largest $d_X^l$ positive eigenvalues,
    and reconstruct the matrix.

    \item \textbf{Cross-request non-negativity projection:}
    for every $(p,q)\in\mathcal{C}^l$, set
    \begin{equation}
    G_{pq}
    \leftarrow
    \max(G_{pq},0),
    \qquad
    G_{qp}
    \leftarrow
    G_{pq}.
    \end{equation}

    \item \textbf{Within-request identity projection:}
    for each request $t$, set
    \begin{equation}
    G[\mathcal{I}_t^l,\mathcal{I}_t^l]
    \leftarrow
    I_{|\mathcal{I}_t^l|}.
    \end{equation}
\end{enumerate}
The procedure is repeated until the maximum number of iterations is
reached or the change in the Gram matrix falls below the prescribed
tolerance:
\begin{equation}
\frac{
\|G^{(k+1)}-G^{(k)}\|_F
}{
\|G^{(k)}\|_F+\epsilon
}
<\eta.
\label{eq:supp_gram_stopping}
\end{equation}

After convergence, we eigendecompose the projected Gram matrix:
\begin{equation}
\widehat G_X^l
=
E_X^l
\Lambda_X^l
(E_X^l)^\top.
\label{eq:supp_gram_eigendecomposition}
\end{equation}
Let $r_X^l=\operatorname{rank}(\widehat G_X^l)$. A compact factor realizing this Gram geometry is
\begin{equation}
Z_X^l
=
(\Lambda_{X,+}^l)^{1/2}
(E_{X,+}^l)^\top
\in
\mathbb{R}^{r_X^l\times N_l},
\qquad
(Z_X^l)^\top Z_X^l
=
\widehat G_X^l,
\label{eq:supp_gram_factor}
\end{equation}
where $N_l$ is the total number of retained directions and the $+$ subscript denotes the positive eigenspace. When $r_X^l<d_X^l$, $Z_X^l$ is padded with zero rows to match the ambient direction dimension. Because a Gram matrix determines its factor only up to a left orthogonal transformation, the recovered directions are aligned with the original concatenated directions using an orthogonal Procrustes step before being partitioned back into request-specific blocks.
The resulting left and right blocks are combined with the retained
singular values to reconstruct the reconfigured core updates.

Cyclic projection over a non-convex feasible set does not guarantee a
global optimum.
We therefore interpret this procedure as an approximate solver that
seeks a nearby feasible geometry.
In practice, the proximity objective and deterministic initialization
provide stable solutions across the evaluated request scales.

\subsection{Limitations and Scope}
\label{sec:supp_limitations}

MCU provides an efficient merging-based solution to continual
multimodal unlearning, but several limitations remain.

\paragraph{Conservative compatibility surrogate.}
The separate non-negativity constraints on the left and right Gram
matrices are sufficient but not necessary for a non-antagonistic
rank-one interaction.
They may therefore modify some direction pairs whose sign-invariant
product is already non-negative.
The proximity objective limits this effect, but it does not make the
surrogate exact.

\paragraph{Parameter geometry versus functional behavior.}
The proposed interaction measures characterize relationships among
parameter-space update directions.
They do not fully determine how two updates interact on model outputs,
hidden representations, or gradients.
The leave-one-task-out analysis provides behavioral evidence that the
geometric reconfiguration reduces cross-task deviations, but a
complete functional characterization remains an open direction.

\paragraph{Basis dependence of capacity control.}
Row norms are defined in the deterministic shared core basis and are
not invariant under arbitrary rotations of that basis.
Accordingly, RowCap should be interpreted as controlling
core-coordinate concentration under the chosen representation rather
than as identifying basis-free model channels.

\paragraph{Dependence on one-shot adapters.}
MCU consolidates, rather than replaces, an underlying one-shot
unlearning method.
Its final performance is therefore constrained by the quality of the
request-specific adapters supplied to the merging stage.
Weak or unstable one-shot unlearning adapters cannot be fully corrected
by geometric processing alone.

\paragraph{Temporary adapter storage.}
Before consolidation, storing the request-specific LoRA bank incurs a
cost that grows linearly with the number of requests.
Although this cost is substantially smaller than archiving
full-model checkpoints under rollback requirements, it is not
constant-memory online unlearning.
Hierarchical consolidation may reduce storage, but can discard
request-level structure.
Our experiments cover two multimodal model backbones and two unlearning benchmarks.
The order-robustness study considers the original, reversed, and one
fixed random permutation.
Broader evaluation across additional architectures, request
distributions, and adversarial relearning settings would further
clarify the generality and security properties of MCU.

\end{document}